\documentclass[10pt,twocolumn,letterpaper]{article}

\usepackage[pagenumbers]{iccv} 

\usepackage{xcolor}
\usepackage{multirow}
\def\deg{$^\circ$}

\usepackage{adjustbox}

\definecolor{iccvblue}{rgb}{0.21,0.49,0.74}
\usepackage[pagebackref,breaklinks,colorlinks,allcolors=iccvblue]{hyperref}

\def\paperID{11602} 
\def\confName{ICCV}
\def\confYear{2025}

\title{IM360: Large-scale Indoor Mapping with 360 Cameras}

\author{Dongki Jung$^{\thanks{These two authors contributed equally.}}$\:\: Jaehoon Choi$^{\footnotemark[1]}$\:\: Yonghan Lee\:\: Dinesh Manocha\\
University of Maryland, College Park
}

\begin{document}
\maketitle

\begin{abstract}

We present a novel 3D mapping pipeline for large-scale indoor environments.
To address the significant challenges in large-scale indoor scenes, such as prevalent occlusions and textureless regions, we propose \it{IM360}, a novel approach that leverages the wide field of view of omnidirectional images and integrates the spherical camera model into the Structure-from-Motion (SfM) pipeline.
Our SfM utilizes dense matching features specifically designed for 360$^\circ$ images, demonstrating superior capability in image registration.
Furthermore, with the aid of mesh-based neural rendering techniques, we introduce a texture optimization method that refines texture maps and accurately captures view-dependent properties by combining diffuse and specular components.
We evaluate our pipeline on large-scale indoor scenes, demonstrating its effectiveness in real-world scenarios.
In practice, IM360 demonstrates superior performance, achieving a 3.5 PSNR increase in textured mesh reconstruction. 
We attain state-of-the-art performance in terms of camera localization and registration on Matterport3D and Stanford2D3D. 
Project page: \url{https://jdk9405.github.io/IM360/}

\end{abstract}



\section{Introduction}
\label{introduction}

\begin{figure}[t]
    \centering
    \begin{subfigure}[t]{1.\linewidth}
    \includegraphics[width=1.\linewidth]{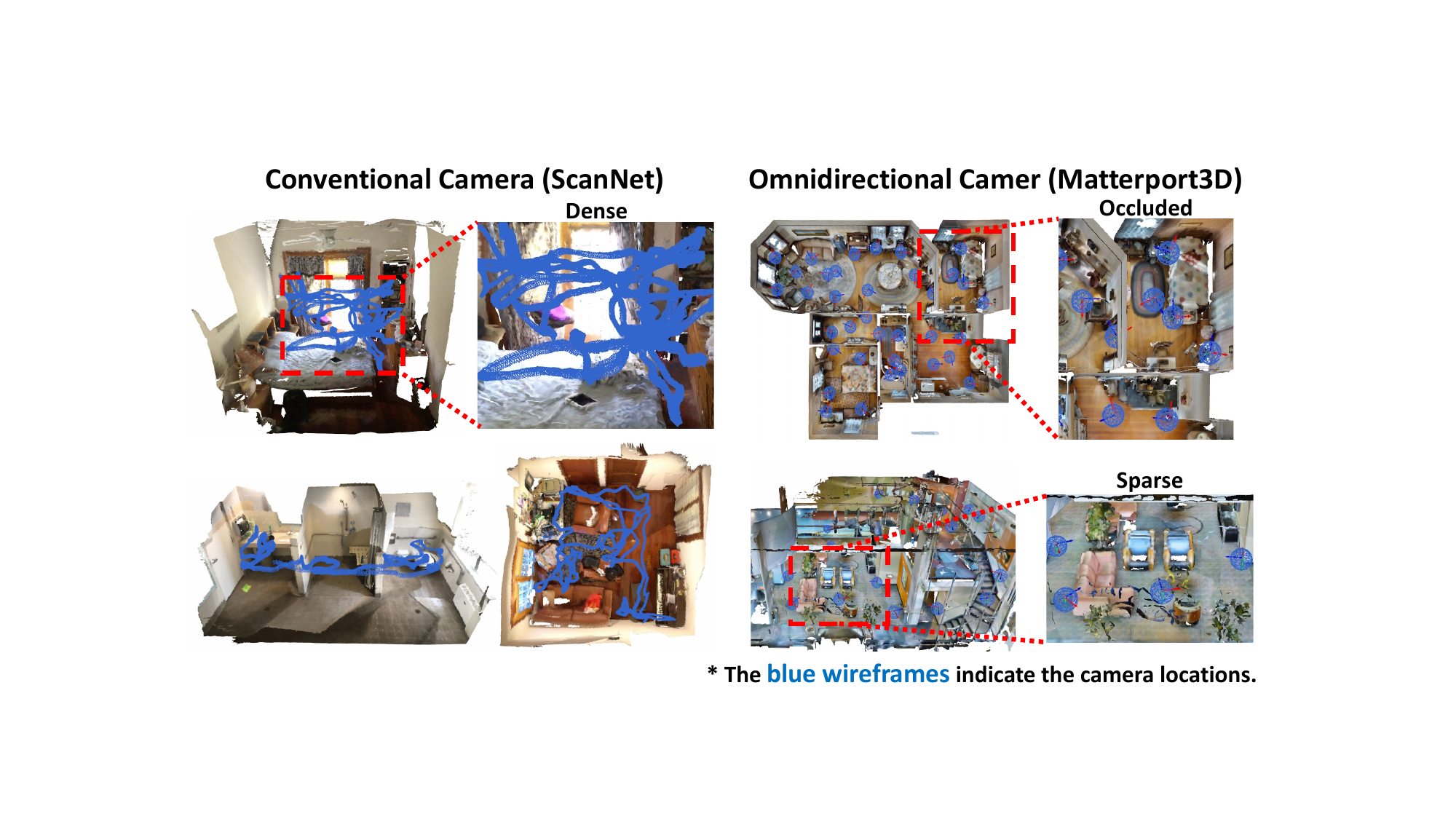}
    \vspace*{-4mm}
    \end{subfigure}


    \centering
    \begin{subfigure}[t]{1.\linewidth}
    \resizebox{\linewidth}{!}{\normalsize
    \begin{tabular}{lcccc}
    \toprule
    Dataset & \# Scenes & \# Images & Floor Space (m\(^2\)) & View Density (images/m\(^2\)) \\ 
    \midrule
    Stanford2D3D \cite{Stanford2d3d}  & 6  & 1,413 (Omnidirectional)  & 6020  & \textcolor{red}{\textbf{0.23}} \\
    Matterport3D \cite{chang2017matterport3d} & 90  & 10,800 (Omnidirectional)  & 46,561 & \textcolor{red}{\textbf{0.23}} \\
    ScanNet \cite{dai2017scannet} & 707 & 2,492,518 (Conventional) & 34,453 & 72.35 \\
    \bottomrule
    \end{tabular}}
    \end{subfigure}

    \caption{
    While the wide field of view of the omnidirectional camera significantly reduces the data acquisition time, it introduces challenges such as sparse views and occlusions. 
    According to the Table, Matterport3D \cite{chang2017matterport3d} and Stanford2D3D \cite{Stanford2d3d} are captured with significantly lower view density compared to ScanNet \cite{dai2017scannet}.
    Our approach provides an effective solution to handle issues in sparsely scanned indoor environments and enables efficient reconstruction and rendering using omnidirectional cameras.
    }
    \vspace{-5mm}
    \label{fig:introduction}
\end{figure}

Indoor 3D mapping and photorealistic rendering are core technologies for various applications in computer vision, robotics, and graphics. High-fidelity digitization of the real world enables immersive experiences in AR/VR and helps bridge the sim-to-real gap in robotic applications. However, the conventional image acquisition process \cite{dai2017scannet,baruch2021arkitscenes,barron2022mip} is often labor-intensive and time-consuming due to the limited field-of-view and sparse scene coverage, as shown in Fig \ref{fig:introduction}. 
Previous studies \cite{janiszewski2022rapid,herban2022use} have demonstrated that omnidirectional cameras can significantly reduce acquisition time, which is crucial for capturing various indoor environments.
Nevertheless, most recent research \cite{schonberger2016structure,moulon2017openmvg,mildenhall2021nerf,kerbl20233d,waechter2014TexRecon} relies on conventional cameras with limited fields of view, which results in inadequate scene coverage, susceptibility to motion blur, and substantial time requirements for capturing large-scale indoor environments. 

Unlike conventional cameras, which benefit from well-established photogrammetric software such as COLMAP \cite{schonberger2016structure} and OpenMVG \cite{moulon2017openmvg}, research on omnidirectional cameras remains fragmented across various domains and lacks a robust pipeline for Structure-from-Motion (SfM) and photorealistic rendering. Notably, there has been limited academic exploration of sparse scanning scenarios aimed at significantly reducing image acquisition time. There are two key challenges in developing 3D mapping and rendering for omnidirectional cameras.

First, the SfM pipeline in sparsely scanned indoor environments suffer from the prevalence of large textureless regions, which degrade the accuracy of both feature detection and matching. Moreover, frequent occlusions in indoor environments, combined with sparse input views as shown in Fig. \ref{fig:introduction}, make the feature matching process even more challenging. In such scenarios, conventional perspective cameras further worsen the problem due to their limited field of view, which reduces the overlap between views. This lack of overlap complicates the estimation of accurate camera poses, making the reconstruction task considerably more difficult. Recently, detector-free or dense matchers \cite{sun2021loftr,melekhov2019dgc,truong2020glu,truong2021learning,edstedt2023dkm,edstedt2024roma} have emerged as a promising alternative to detector-based approaches \cite{detone2018superpoint,revaud2019r2d2,tyszkiewicz2020disk}, particularly in handling repetitive or indiscriminate regions where keypoint detection performance tends to degrade.

Additionally, in sparsely scanned scenarios, recent neural rendering methods \cite{mildenhall2021nerf,kerbl20233d} often face challenges in producing high-quality novel-view images. 
It is widely recognized that both neural radiance fields (NeRF) \cite{mildenhall2021nerf} and 3D Gaussian splatting (3DGS) \cite{kerbl20233d} are optimized for densely captured scenes with substantial overlap between viewpoints.
Prior studies \cite{xiong2023sparsegs,wang2023sparsenerf,martins2024feature,lee2024modegs} have demonstrated that neural rendering algorithms often exhibit overfitting to the training views and a poor rendering quality on novel-view images far from training views.

\paragraph{Main Results}
We present a novel indoor 3D mapping and rendering pipeline designed for omnidirectional images and address the challenges of sparsely scanned indoor environments.
We tackle the problem of extensive textureless regions by employing detector-free matching methods~\cite{EDM}.
To mitigate the frequent occlusions encountered in sparsely scanned images, we utilize an spherical camera model with a wide field of view, increasing the overlap between images and enhancing robustness in multi-view matching.
 While traditional methods, such as OpenMVG \cite{moulon2017openmvg} and COLMAP \cite{schonberger2016structure}, fail to register nearly half of the images in challenging datasets like Matterport3D \cite{chang2017matterport3d} and Stanford2D3D \cite{Stanford2d3d}, our method achieves unparalleled registration performance with high pose accuracy, thanks to the tight integration of a wide-view spherical model.
To the best of our knowledge, ours is the first work to present a complete SfM pipeline dedicated to spherical camera models, where entire step—feature matching, two-view geometry estimation, track triangulation, and bundle adjustment—is conducted on a spherical manifold.

We extend our approach to mesh reconstruction and texturing and provide a comprehensive solution for indoor 3D mapping and rendering.  For mesh reconstruction, we leverage neural surface reconstruction \cite{xiao2024debsdf} by training a signed distance field and extracting a mesh. Classical texture mapping \cite{waechter2014TexRecon} is then combined with texture optimization through differentiable rendering. We optimize textures and train small multi-layer perceptrons (MLPs) to represent specular color. By jointly optimizing diffuse and specular textures along with the MLPs via differentiable rendering \cite{nvdiffrast} under image supervision, our approach achieves superior rendering quality. Additionally, it demonstrates robustness in sparsely scanned indoor environments, outperforming recent neural rendering methods. The proposed method demonstrates significant improvements over traditional approaches, particularly in challenging indoor environments with sparse and textureless scenes. 


We evaluate our approach using the Matterport3D dataset~\cite{chang2017matterport3d}, which offers a larger physical scale than typical indoor datasets like ScanNet, according to~\cite{ramakrishnan2021hm3d}. We also evaluate the performance on
the Stanford2D3D datasets \cite{Stanford2d3d}. The novel contributions of our work include:
\begin{itemize}
    \item We introduce a unified framework for reconstructing textured meshes using omnidirectional cameras in sparsely scanned, large-scale indoor scenes.
    \item We propose the first spherical Structure-from-Motion approach utilizing dense matching features specifically designed for omnidirectional cameras.
    \item We integrate classical texture mapping with neural texture fine-tuning through differentiable rendering, demonstrating improved rendering quality, in sparsely scanned, in-the-wild indoor environments. 
\end{itemize}
We compare our method with SOTA and observe higher accuracy in terms of camera localization and rendering. In particular our approach can render higher frequency details and results in lower noise in the reconstructed models.

\section{Related Work}
\subsection{Structure from Motion}

Structure from Motion (SfM) is a fundamental problem in computer vision. Traditional SfM frameworks typically begin with the detection of keypoints and descriptors, which serve as a core component, ranging from classical algorithms \cite{SIFT} to learning-based techniques \cite{detone2018superpoint,sarlin2020superglue,revaud2019r2d2}. Then, they match these keypoints by either nearest neighbor \cite{schonberger2016structure} or learning-based methods \cite{sarlin2020superglue}. The matching pair is verified by recovering the two-view geometry \cite{hartley2003multiple} with RANSAC \cite{RANSAC}. Based on these matching pairs, popular SfM methods \cite{schonberger2016structure,moulon2017openmvg,sweeney2015theia} follow incremental approach, sequentially registering new images and reconstructing their 3D structures through triangulation \cite{hartley2003multiple} and bundle adjustment \cite{Bundleadjustment}. Alternatively, a recent work \cite{GLOMAP} employs a global approach, recovering the camera geometry for all input images simultaneously. The 2D correspondence matching significantly impact the overall performance of SfM pipeline. In textureless region of indoor environment, these methods suffer from poor feature detection \cite{he2024detector}. Detector-free or dense matching methods  \cite{sun2021loftr,edstedt2023dkm,edstedt2024roma} is proposed to solve this issue by estimating dense feature matches at pixel level. DetectorFreeSfM \cite{he2024detector} builds these detector-free matches \cite{sun2021loftr} and refines the tracks and geometry of
the coarse SfM model by enforcing multi-view consistency. 

Conversely, there are relatively few open-source software packages \cite{moulon2017openmvg,jiang20243d} that support omnidirectional cameras. 
These frameworks typically employ an incremental SfM workflow based on a spherical camera model \cite{jiang20243d}. 
However, existing methods lack a comprehensive pipeline specifically designed for omnidirectional images. 
Recently, the first dense matching method for omnidirectional images, EDM \cite{EDM}, has been introduced. 
Building upon this feature matching and two-view geometry estimation \cite{solarte2021robust}, we propose a complete pipeline for spherical SfM to address the challenges inherent in indoor scene reconstruction.

\subsection{Textured Mesh Reconstruction}
Textured mesh is the basic component of photorealistic rendering. Traditional computer vision has developed effective techniques for mapping between texture and geometry. Classical texture mapping \cite{lempitsky2007seamless,waechter2014TexRecon,fu2018texture} employs a Markov Random Field to select the optimal color image for each mesh face and applies global color adjustment for consistency \cite{waechter2014TexRecon}. Recently, numerous neural rendering techniques have been developed for surface reconstruction \cite{wang2021neus,yu2022monosdf,choi2023tmo,xiao2024debsdf} and rendering \cite{mildenhall2021nerf,barron2022mip,barron2023zip,kerbl20233d}. Notably, some approaches leverage differentiable rendering \cite{nvdiffrast,PyTorch3D,Mitsuba3,munkberg2021nvdiffrec,Goel_2022_CVPR} to learn neural representations of texture, lighting, and geometry. Additionally, certain NeRF variants \cite{SNeRG,chen2023mobilenerf,NeuRas,DNMP,Choi2024LTM} employ differentiable volumetric rendering with mesh primitives, which is similar to rasterization-based rendering. Some neural rendering methods are tailored for omnidirectional cameras. EgoNeRF \cite{choi2023balanced} utilizes a spherical feature grid to represent a distant environment for rendering. OmniSDF \cite{kim2024omnisdf} introduces an adaptive spherical binoctree method for surface reconstruction. However, in sparsely scanned indoor environment, which contains lots of textureless regions and sparse viewpoints, most neural rendering methods struggle with severe artifacts. To address this limitation, we propose a hybrid approach that combines classical texture mapping with neural texture optimization.


\section{Our Method}

\begin{figure}[t]
    \centering
    \includegraphics[width=1.0\linewidth]{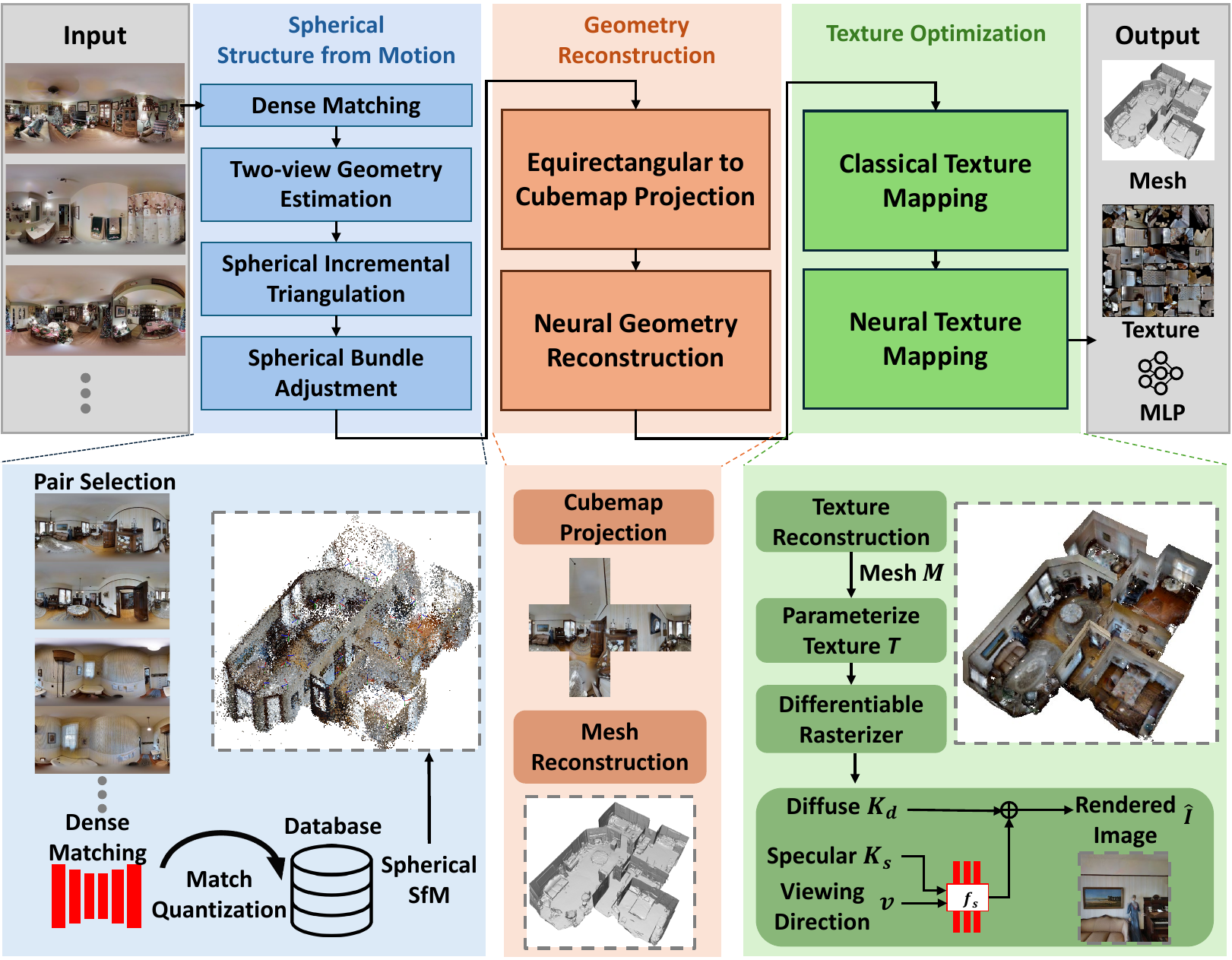}
    \caption{
    \textbf{Overview of Our Pipeline}. It is comprised of three steps: Spherical Structure from Motion (Sec. \ref{sec:3_1}), Geometry Reconstruction (Sec. \ref{sec:3_2}), and Texture Optimization (Sec. \ref{sec:3_3}).
    In Section \ref{sec:3_1}, spherical SfM is utilized to estimate the camera poses for omnidirectional images in challenging indoor environments.
    Then, in Section \ref{sec:3_2}, our method apply cubemap projection and estimate monocular depth and normal to train geometry $f_{sdf}$ representation using volumetric rendering. In Section \ref{sec:3_3}, we initlize texture using classical texture mapping and jointly optimize diffuse, specular texture along with small MLP for modeling appearance.}
     \vspace*{-2mm}
     \label{fig:pipeline}
\end{figure}

Given a set of equirectangular projection (ERP) images, our goal is to reconstruct the geometry and texture of real-world scenes. In Fig. \ref{fig:pipeline}, 
our pipeline is composed of three primary components: Spherical Structure from Motion (SfM), geometry reconstruction, and texture optimization. To address challenges associated with limited overlap and frequent occlusions in indoor environments, our method leverages the wide field of view inherent in spherical imagery. 
We perform SfM utilizing a spherical camera model and dense features to estimate camera poses from sparsely scanned images in textureless regions. 
However, since a naive combination of SfM and dense feature matching is not directly feasible, we refine this approach to ensure compatibility.
For 3D geometric reconstruction accurately, we converted ERP images into cubemaps and conducted training on the resulting perspective images. Then, we employ a neural model \cite{xiao2024debsdf} that represents a signed distance function (SDF). After training the model, we extract the mesh using the Marching Cubes algorithm \cite{marchingcube} and apply classical texture mapping \cite{waechter2014TexRecon} to initialize texture. Finally, we optimize the texture for diffuse, specular and small MLP components using differentiable rendering \cite{nvdiffrast}.    

\subsection{Dense Matching and Structure from Motion}
\label{sec:3_1}

Large-scale indoor scenes often pose significant challenges for camera localization using conventional SfM methods such as COLMAP \cite{schonberger2016structure} or OpenMVG \cite{moulon2017openmvg}, due to numerous occlusions and textureless regions.
To effectively reconstruct sparsely captured indoor scenes, we integrate a spherical camera model into every step of SfM pipeline, combining spherical dense matching, two-view geometry estimation, and optimization methods into a unified, comprehensive 3D reconstruction framework.
We construct our pipeline based on the unit bearing vector $u \in \mathbb{S}^2$,
\begin{equation} \label{plane5}
    u = \left( \sin(\theta)\cos(\phi), \ \sin(\phi), \ \cos(\theta)\cos(\phi) \right)
\end{equation}
where $\theta \in [-\pi, \pi]$ and $\phi \in [-\frac{\pi}{2}, \frac{\pi}{2}]$.
\vspace*{2mm}

\noindent\textbf{Spherical Dense Matching}
Performing spherical SfM requires feature matching as a preliminary step. 
Several feature matching methods specifically designed for 360\deg images have been proposed, such as SPHORB \cite{zhao2015sphorb}, SphereGlue \cite{gava2023sphereglue}, and EDM \cite{EDM}. 
Alternatively, feature matching can also be performed independently on each face of a cubemap generated from a 360\deg image.
Among these approaches, dense feature matching \cite{EDM} recently demonstrated superior feature matching performance for 360° images, especially in sparse scenes with large viewpoint changes.
However, since dense matching methods do not explicitly provide feature descriptors, constructing feature tracks of multiple views for SfM becomes challenging.
Inspired by \cite{sarlin2019coarse, he2024detector}, we apply the additional strategies for match quantization by rounding \cite{chen2022aspanformer} and merging \cite{shen2022semi} techniques, to ensure multiview match consistency.

\noindent\textbf{Spherical Two-view Geometry Estimation}
It can be formally proven that the traditional two-view epipolar constratint for normalized coordinates extends to unit bearing vectors in spherical images \cite{scaramuzza2011visual},
\begin{equation} \label{eq:essentail}
u_1^T  E  u_2 = 0 \; \text{with} \; E = R [t]_\times,
\end{equation}
where $E \in \mathbb{R}^{3\times3}$ is the essential matrix, which can be decomposed into the camera rotation $R \in \text{SO(3)}$ and the translation $t \in \mathbb{R}^3$. 
%
%
Solarte \textit{et al.} \cite{solarte2021robust} employed the 8-point algorithm to solve spherical two-view geometry using the Direct Linear Transform (DLT) framework \cite{hartley2003multiple}.
We adopt this two-view geometry estimation method with additional normalization \cite{solarte2021robust} to enhance numerical stability during Singular Value Decomposition (SVD) computation.
Using those initial pose with rotation $R$ and translation $t$, we activate incremental triangulation and bundle adjustment under the sphere camera model.

\noindent\textbf{Spherical Bundle Adjustment}
By representing each feature track observation as a bearing vector $u$, conventional Bundle Adjustment (BA) can be reformulated in a manner similar to the perspective case, as follows:
\begin{equation} \label{eq:textureupdate}
    L = \sum_{i}\sum_{j}\rho(|| \Pi (P_j; R_i, t_i) - u_{ij}||^2),
\end{equation}
where $\rho$ denotes the robust loss with a soft L1 function in local bundle adjustment, while global bundle adjustment is performed without this robustifier. 
$\Pi( \cdot )$ represents the spherical projection $\Pi: P \rightarrow u$, $\mathbb{R}^3 \mapsto \mathbb{S}^2$, which maps the 3D points $X_j$ from triangulation to unit bearing vectors, given the camera poses $R_j \in \text{SO(3)}$ and $t_j \in \mathbb{R}^3$.

\subsection{Geometric Reconstruction}
\label{sec:3_2}

\begin{figure}[t]
    \centering
    \includegraphics[width=1.0\linewidth]{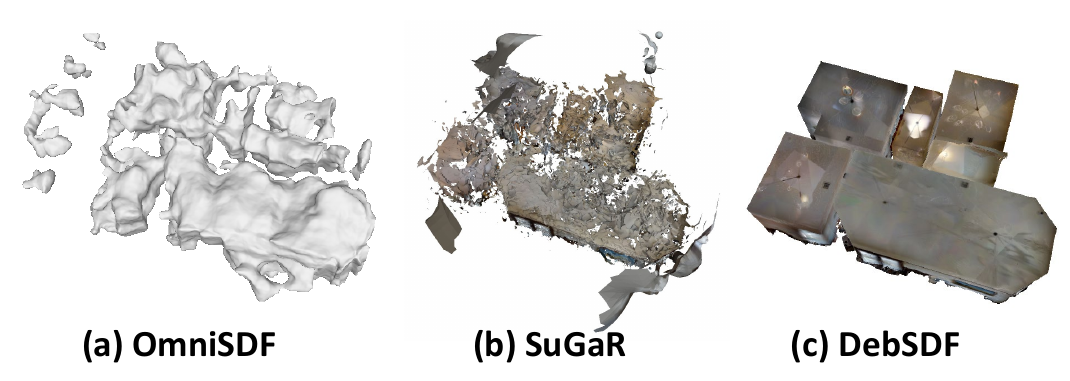}
    \vspace{-7mm}
    \caption{
    Our method, which applies cubemap projection and trains neural surface reconstruction using multiple perspective images, achieves highly accurate geometry reconstruction. In contrast, neural surface reconstruction designed for ERP images (OmniSDF \cite{kim2024omnisdf}) and Gaussian Splatting-based surface reconstruction (SuGaR \cite{guedon2024sugar} and VCR-GauS \cite{chen2025vcr}) fail to generate a clean mesh.    
    }
     \label{fig:geo_recon}
     \vspace{-5mm}
\end{figure}

To address visual localization and mapping in textureless and highly occluded sparsely scanned indoor scenes, we use neural rendering techniques \cite{yu2022monosdf, xiao2024debsdf} for surface reconstruction instead of traditional 3D reconstruction methods such as multi-view stereo (MVS)~\cite{schonberger2016pixelwise}.
Many methods leverage monocular geometric priors~\cite{yu2022monosdf, xiao2024debsdf} and demonstrate higher robustness in textureless and sparsely covered areas. However, 
in terms of ERP images there are no  robust methods for zero-shot monocular depth and normal estimation.
As shown in Fig. \ref{fig:geo_recon}, we also observed  that OmniSDF \cite{kim2024omnisdf}, which learns signed distance functions (SDF) from equirectangular projection (ERP) images, failed to converge when applied to large-scale indoor datasets. 
Despite incorporating geometric priors \cite{eftekhar2021omnidata}, Gaussian Splatting-based surface reconstruction \cite{guedon2024sugar,chen2025vcr} still fails to produce reasonable geometry, resulting in a polygonal soup.

To address these limitations, we first project ERP images onto a cubemap representation and then apply a volumetric rendering approach to estimate the signed distance function (SDF).
Since most existing mesh rendering techniques---except for volumetric methods leveraging monocular geometric priors \cite{yu2022monosdf, xiao2024debsdf}---struggle in sparse indoor scenes,
we employ DebSDF \cite{xiao2024debsdf} for geometric mesh reconstruction and integrate the Marching Cubes algorithm \cite{marchingcube} to generate a high-fidelity mesh.

\subsection{Texture Optimization}
\label{sec:3_3}
In this stage, we project textures onto the 3D mesh $M$ generated in the preceding step. In comparison to recent neural rendering methods such as NeRF \cite{mildenhall2021nerf} and 3DGS \cite{kerbl20233d}, it is important to highlight that texture mapping provides a more robust solution for novel view synthesis in sparsely scanned indoor environments.
Utilizing images obtained from cubemap projections along with their corresponding camera poses, we initially employ TexRecon \cite{waechter2014TexRecon} to construct the texture map $T$. However, this conventional texture mapping approach suffers from visible seams between texture patches and is prone to geometric inaccuracies.

Inspired by TMO \cite{choi2023tmo}, we parameterize the texture map and leverage a differentiable rasterizer $\mathcal{R}$ \cite{nvdiffrast} to render an image from a given viewpoint $P$. 
While TMO \cite{choi2023tmo} is constrained to representing only diffuse textures and fails to capture view-dependent effects, our approach surpasses these limitations by enabling a more comprehensive and physically accurate representation.
The diffuse texture $K_{d} \in \mathbb{R}^3$ is directly converted into an RGB image. Additionally, we initialize a specular feature $K_{s} \in \mathbb{R}^3$ and employ a small MLP $f_{s}$ as a fragment shader. This MLP takes the specular feature $K_{s}$ and viewing direction $v$ as inputs to compute the specular color as follows: 
\begin{equation} \label{eq:textureupdate}
    \hat{I_{d}} = \mathcal{R}(M, K_{d}, P), ~   \hat{I_{s}} = f_{s}(\mathcal{R}(M, K_{s}, P), v) 
\end{equation}
The final rendered color $\hat{I} \in \mathbb{R}^3$ is obtained by combining the diffuse and specular components, $\hat{I} = \hat{I_{d}} + \hat{I_{s}}$. 
To optimize the diffuse, specular features, and an MLP, we employ a combination of L1 and SSIM \cite{SSIM} as the photometric loss in image space. This loss is calculated between the final rendered image and the ground truth image from the corresponding viewpoint, as follows:
\begin{equation} \label{eq:textureupdate1}
    L_{photo} = (1-\alpha)\parallel\hat{I} - I\parallel + \alpha\:(1 - SSIM(\hat{I}, I))
\end{equation}
where $\alpha$ is set to 0.2 and the weight to balance the losses.


\section{Implementation and Performance}
\subsection{Experiment Settings}
\paragraph{Matterport3D Dataset}
The Matterport3D dataset \cite{chang2017matterport3d} consists of 90 indoor scenes encompassing a total of 10,800 panoramic images. Following the EDM approach \cite{EDM}, we define the ground truth camera poses of these images. From the validation and test sets of the official benchmark split, we selected three scenes each, ensuring that these scenes are entirely indoors or composed of levels no higher than the second floor.
Since dense matching requires overlapped image pairs as input,
we assume that during 360-degree camera scanning, the information about which room each camera belongs to is known.
To utilize this information effectively, we leveraged the $house\_segmentation$ labels available in the Matterport3D dataset. Images taken within the same room were selected as matching pairs, while for transitions between different rooms, only one image pair with the highest overlap was linked.
For EDM \cite{EDM}, the ERP images were resized to $640 \times 320$, whereas images for other feature matching methods were resized to $1024 \times 512$.

\paragraph{Stanford2D3D Datatset}
The Stanford2D3D dataset \cite{Stanford2d3d} comprises scenes from six large-scale indoor environments collected from three different buildings. 
We utilize this dataset to demonstrate that our spherical SfM method outperforms existing SfM approaches in terms of registration performance and accurate pose estimation. 
Specifically, we selected three scenes containing a moderate number of images. 
Assuming that overlapping image pairs were predefined during the scanning process, we determined covisibility and defined matching pairs using geometric meshes and ground truth poses. 
Additionally, we resized the equirectangular projection (ERP) images following the same methodology employed in the Matterport3D dataset.

\paragraph{Implementation Details}
We construct the Spherical Structure from Motion based on SphereSfM \cite{jiang20243d} and COLMAP \cite{schonberger2016structure}. To address challenges in textureless or occluded regions, we leverage the dense matching algorithm EDM \cite{EDM} for enhanced feature extraction. Following the DebSDF \cite{xiao2024debsdf}, the SDF and color networks are implemented using an 8-layer MLP and a 2-layer MLP, respectively, each with a hidden dimension of 256. The networks are optimized for geometric reconstruction using the Adam optimizer \cite{kingma2014adam}, starting with an initial learning rate of $5 \times 10^{-4}$ and applying exponential decay at each iteration. Input images are resized to $384 \times 384$, and geometric cues are provided by the pretrained Omnidata model \cite{eftekhar2021omnidata}. In Texture Optimization, the specular features are initialized to zero values, matching the resolution of the diffuse texture image. The small MLP consists of 2 layers with 32 hidden dimensions. We employ the Adam optimizer \cite{kingma2014adam} with a learning rate of $5 \times 10^{-4}$ and utilize a learning rate scheduler. The model is trained for 7,000 iterations. We render images at a resolution of $512 \times 512$.


\begin{figure}[t]
    \centering    
    \includegraphics[width=1.0\linewidth]{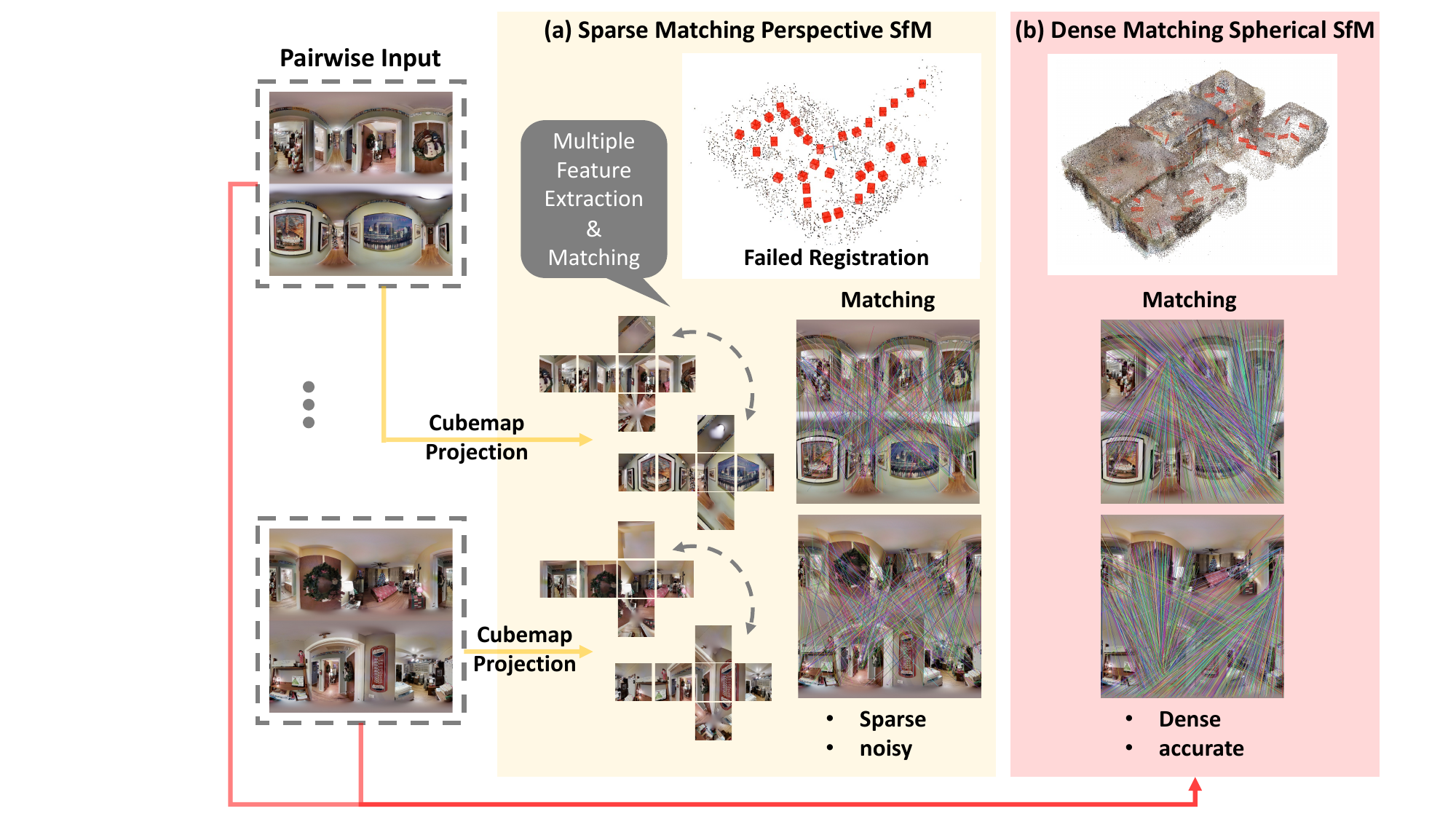}
    \caption{
    We provide a visual comparison between sparse matching perspective SfM and our proposed dense matching spherical SfM. 
    In (a), ERP images are converted into a cubemap representation, after which feature matching is performed across all 36 possible image pairs, resulting in sparse and noisy correspondence matches. (b) demonstrates our approach, which directly finds dense and accurate correspondences on ERP images, thereby facilitating the construction of a detailed 3D structure.
    }
    \label{fig:sfm}
\end{figure}

\subsection{Experimental Results}
\paragraph{Camera Pose Estimation}
Accurate geometric 3D reconstruction depends on precise camera pose estimation. 
We compare our proposed method, which leverages dense matching \cite{EDM} and Spherical SfM, against four different methods:
1) \textbf{OpenMVG:} An open-source SfM pipeline that supports spherical camera models \cite{moulon2017openmvg}.
2) \textbf{SPSG COLMAP:} This approach employs SuperPoint \cite{detone2018superpoint} and SuperGlue \cite{sarlin2020superglue} for feature detection and matching on perspective images. ERP images are transformed into a cubemap representation, yielding six perspective views per ERP image. Feature matching is performed in a brute-force manner across all 36 possible image pairs. Subsequently, incremental triangulation and bundle adjustment are performed under the rig constraints for each 360 camera node.
3) \textbf{DKM COLMAP:} This method leverages DKM \cite{edstedt2023dkm} to establish dense correspondences without relying on explicit feature extraction. Since DKM is designed for perspective images, it also follows the cubemap projection and exhaustive pairwise matching, and rig-based optimization.
4) \textbf{SphereGlue COLMAP:} Building on COLMAP, SphereSfM \cite{jiang20243d} incorporates spherical camera models and spherical bundle adjustment to handle ERP images. For feature extraction and matching, SuperPoint \cite{detone2018superpoint} with a local planar approximation \cite{eder2020tangent} and SphereGlue \cite{gava2023sphereglue} are utilized to mitigate distortion in ERP images.
To estimate initial poses for incremental triangulation, we follow the normalization strategy and non-linear optimization proposed by Solarte et al. \cite{solarte2021robust}.

\begin{figure}[t]
    \centering
    \includegraphics[width=1.0\linewidth]{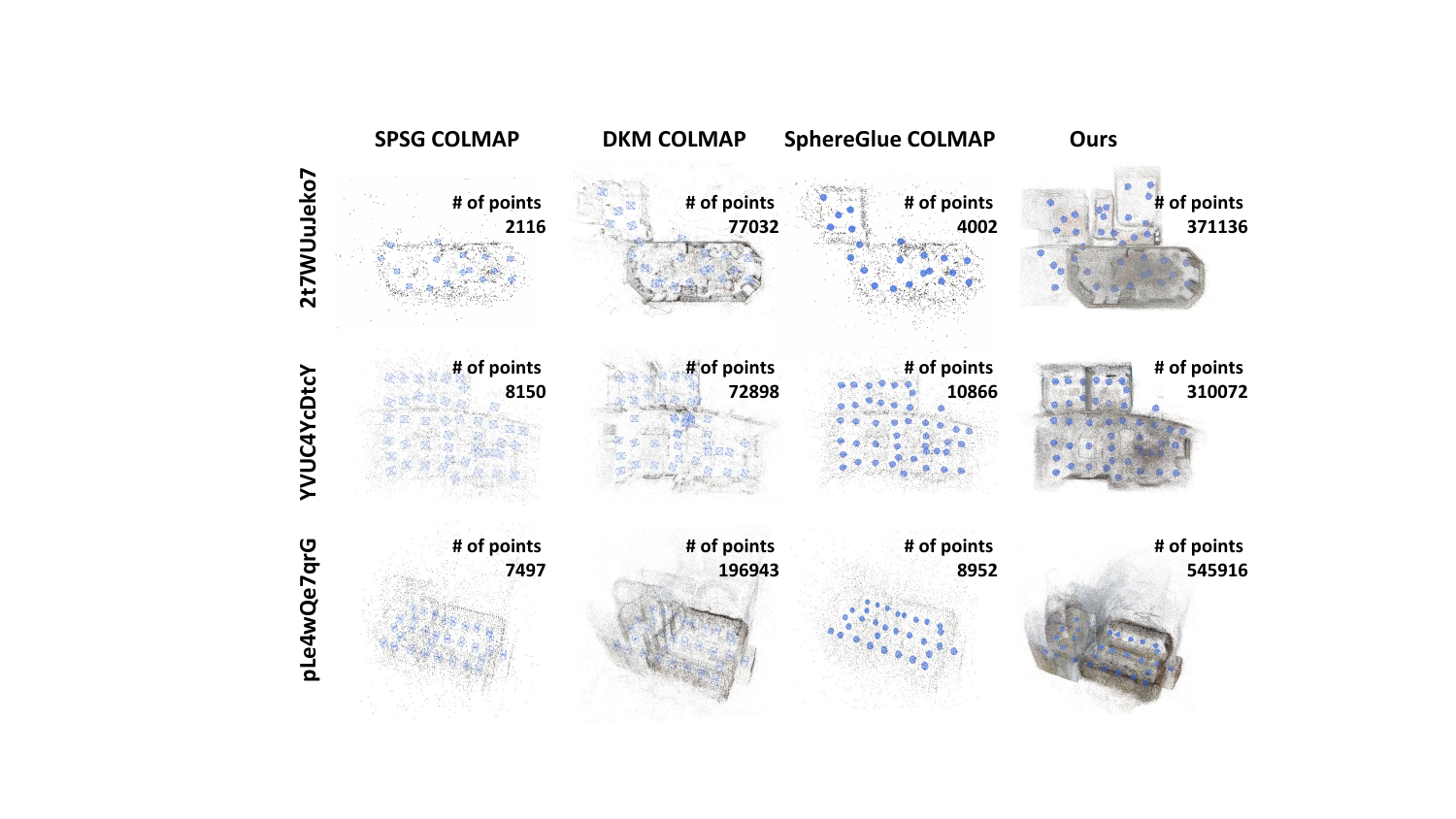}
    \caption{\textbf{Qualitative Comparison of SfM results on Matterport3D.} By leveraging dense features on equirectangular projection (ERP) images, our method effectively finds correspondences in textureless regions. }
    \vspace*{-2mm}
    \label{fig:sfm_result}
\end{figure}

Table \ref{table:mp3d} presents the quantitative results of camera pose estimation on the Matterport3D dataset. 
While previous methods also performed well in smaller scenes with significant image overlap, our proposed method demonstrated superior robustness across diverse scenes. 
By leveraging a wide field of view (FoV) for matching, our approach proved particularly effective on datasets with sparse input views. 
Notably, it achieved the most accurate registrations in scenes with frequent occlusions. 
Figure \ref{fig:sfm_result} visualizes the estimated camera poses and triangulated 3D points, demonstrating that our method estimated both the largest number of 3D points and the most camera poses.

\begin{table*}[t]
    \begin{center}
    \resizebox{0.95\linewidth}{!}{
    \begin{tabular}{l|cccc cccc cccc}
        \toprule
        \multirow{2}{*}{Method} & \multicolumn{4}{c}{2t7WUuJeko7} & \multicolumn{4}{c}{8194nk5LbLH} & \multicolumn{4}{c}{pLe4wQe7qrG} \\
        \cmidrule{2-5}\cmidrule(lr){6-9}\cmidrule(lr){10-13}
        & \# Registered & AUC @3\textdegree & AUC @5\textdegree & AUC @10\textdegree & \# Registered & AUC @3\textdegree & AUC @5\textdegree & AUC @10\textdegree & \# Registered & AUC @3\textdegree & AUC @5\textdegree & AUC @10\textdegree \\
        \midrule
        OpenMVG & 5 / 37 & 0.82 & 1.10 & 1.30 &2 / 20 & 0.27&0.37 &0.45 &25 / 31 & 45.11 &52.82 &58.67 \\
        SPSG COLMAP & 16 / 37 & 14.30 & 15.79 & 16.90 & 12 / 20 & 26.64 & 29.88 & 32.31 & \textbf{31 / 31} & 75.02 & 85.01 & 92.51 \\
        DKM COLMAP & 22 / 37& 25.14 & 27.70 & 29.62 & 9 / 20 & 14.64 & 16.36 & 17.66 & \textbf{31 / 31} & \textbf{75.94} & \textbf{85.56} & \textbf{92.78} \\
        SphereGlue COLMAP & 21 / 37 & 23.95 & 26.98 & 29.26 & 12 / 20 & 23.11  & 27.76 & 31.24 & \textbf{31 / 31} & 66.83 & 80.06 & 90.03 \\
        IM360 (Ours) & \textbf{37 / 37} & \textbf{49.16} & \textbf{69.05} & \textbf{84.53} & \textbf{20 / 20} & \textbf{34.91} & \textbf{44.16} & \textbf{66.87} & \textbf{31 / 31} & 73.95 & 84.37 & 92.18 \\
        \midrule
        \multirow{2}{*}{Method} & \multicolumn{4}{c}{RPmz2sHmrrY} & \multicolumn{4}{c}{YVUC4YcDtcY} & \multicolumn{4}{c}{zsNo4HB9uLZ} \\
        \cmidrule{2-5}\cmidrule(lr){6-9}\cmidrule(lr){10-13}
        & \# Registered & AUC @3\textdegree & AUC @5\textdegree & AUC @10\textdegree & \# Registered & AUC @3\textdegree & AUC @5\textdegree & AUC @10\textdegree & \# Registered & AUC @3\textdegree & AUC @5\textdegree & AUC @10\textdegree \\
        \midrule
        OpenMVG & 22 / 59 & 9.86 &11.31 &12.41 &8 / 46 &2.22 &2.41 &2.56 &2 / 53 &-- &-- &-- \\
        SPSG COLMAP & 34 / 59 & 22.93 & 26.87 & 29.83 & \textbf{46 / 46} & \textbf{84.21} & \textbf{90.53} & \textbf{95.26} & 19 / 53 & 7.84 & 9.22 & 10.55 \\
        DKM COLMAP & 51 / 59 & 41.56 & 53.56 & 62.58 & \textbf{46 / 46} & 56.33 & 61.00 & 64.67 & 19 / 53 & 7.87 & 8.71 & 9.41 \\
        SphereGlue COLMAP & 34 / 59 & 22.23 & 26.45 & 29.62 & \textbf{46 / 46} & 78.32 & 86.97 & 93.48 & 22 / 53 & 6.25 & 7.56 & 9.46 \\
        IM360 (Ours) & \textbf{59 / 59} & \textbf{53.44} & \textbf{71.87} & \textbf{85.93} & \textbf{46 / 46} & 72.40 & 83.40 & 91.71 & \textbf{53 / 53} & \textbf{52.35} & \textbf{71.14} & \textbf{85.49} \\
        \bottomrule
    \end{tabular}
    }
    \end{center}
    \vspace*{-4mm}
    \caption{\label{table:mp3d} \textbf{Quantitative Evaluation of Camera Localization Performance} on the Matterport3D dataset \cite{chang2017matterport3d}. Our method demonstrates superior results in terms of registration accuracy and achieves comparable performance in pose estimation across most scenes.}
\end{table*}

\begin{table*}[t]
    \begin{center}
    \resizebox{0.95\linewidth}{!}{
    \begin{tabular}{l|cccc cccc cccc}
        \toprule
        \multirow{2}{*}{Method} & \multicolumn{4}{c}{area 3} & \multicolumn{4}{c}{area 4} & \multicolumn{4}{c}{area 5a} \\
        \cmidrule{2-5}\cmidrule(lr){6-9}\cmidrule(lr){10-13}
        & \# Registered & AUC @3\textdegree & AUC @5\textdegree & AUC @10\textdegree & \# Registered & AUC @3\textdegree & AUC @5\textdegree & AUC @10\textdegree & \# Registered & AUC @3\textdegree & AUC @5\textdegree & AUC @10\textdegree \\
        \midrule
        OpenMVG & 6 / 85 & 0.35 &0.38 &0.40 & 17 / 258 &0.34 &0.37 &0.39 &  8 / 143 &0.23 &0.25 &0.26 \\
        SPSG COLMAP & 28 / 85 & 6.71 & 7.47 & 8.29 & 73 / 258 & 4.75 & 5.77 & 6.55 & 54 / 143 & 8.84 & 10.93 & 12.51 \\
        DKM COLMAP & 45 / 85& 13.10 & 14.92 & 16.32 & 79 / 258 & 3.39 & 3.96 & 4.40 & 78 / 143 & 15.27 & 18.71 & 21.38  \\
        SphereGlue COLMAP & 30 / 85 & 5.87 & 7.38 & 8.75 & 116 / 258 & 8.62  & 11.64 & 14.13 & 69 / 143 & 5.10 & 8.06 &  11.73\\
        IM360 (Ours) & \textbf{85 / 85} & \textbf{32.20} & \textbf{45.12} & \textbf{58.23} & \textbf{258 / 258} & \textbf{28.70} & \textbf{50.50} & \textbf{71.29} & \textbf{138 / 143} & \textbf{19.52} & \textbf{32.54} & \textbf{51.72}  \\
        \bottomrule
    \end{tabular}
    }
    \end{center}
    \vspace*{-4mm}
    \caption{\label{table:stfd} \textbf{Quantitative evaluation of camera Localization Performance} on the Stanford2D3D dataset \cite{Stanford2d3d}. Our method outperforms all other approaches in terms of registration accuracy and pose estimation.}
    \vspace*{-2mm}
\end{table*}

\begin{table*}[t]
    \centering
    \begin{adjustbox}{width=0.94\linewidth,center}
    \begin{tabular}{l|c|ccccccc}
    \toprule
    \multirow{2}{*}{Method}&\multirow{2}{*}{Rendering}&\multicolumn{7}{c}{PSNR \(\uparrow\)  / SSIM \(\uparrow\)  / LPIPS \(\downarrow\) } \\
    & & 2t7WUuJeko7 & 8194nk5LbLH & pLe4wQe7qrG & RPmz2sHmrrY & YVUC4YcDtcY & zsNo4HB9uLZ & Mean \\
    \midrule
    ZipNeRF \cite{barron2023zip} & Volume &  15.1 / 0.51 / 0.69 & 11.9 / 0.44 / 0.74 & 14.1 / 0.43 / 0.70 & 13.1 / 0.50 / 0.69 & 15.3 / 0.53 / 0.68 & 14.1 / 0.65 / 0.68 & 13.9 / 0.51 / 0.68 \\
    3DGS \cite{kerbl20233d} & Splat & 14.4 / 0.47 / 0.53 & 11.9 / 0.38 / 0.60 & 13.5 / 0.37 / 0.56 & 12.9 / 0.51 / 0.55 & 14.2 / 0.49 / 0.52 & 13.7 / 0.60 / 0.50 & 13.4 / 0.47 / 0.55 \\
    SparseGS \cite{xiong2023sparsegs} & Splat & 15.4 / 0.48 / 0.50 & 12.8 / 0.36 / 0.58 & 13.8 / 0.37 / 0.52 & 13.6 / 0.46 / 0.57 & 15.5 / 0.50 / 0.49 & 14.4 / 0.58 / 0.55 & 14.3 / 0.46 / 0.53 \\
    TexRecon \cite{waechter2014TexRecon} & Mesh & 17.1 / 0.51 / 0.43 & 14.2 / 0.47 / 0.49 & 15.9 / 0.45 / 0.46 & 16.7 / 0.60 / 0.38 & 15.0 / 0.59 / 0.42 & 17.0 / 0.51 / 0.42 & 15.9 / 0.54 / 0.43 \\
    \midrule
    IM360* (Ours) & Mesh & 19.6 / 0.60 / 0.37 & 16.7 / 0.60 / 0.43 & 18.5 / 0.65 / 0.36 & 18.9 / 0.65 / 0.38 & 18.3 / 0.60 / 0.42 & 19.8 / 0.68 / 0.41 & 18.6 / 0.63 / 0.39 \\
    IM360 (Ours) & Mesh & \textbf{19.8} / \textbf{0.61} / \textbf{0.38} & \textbf{16.9} / \textbf{0.61} / \textbf{0.43} & \textbf{18.8} / \textbf{0.66} / \textbf{0.35} & \textbf{20.6} / \textbf{0.75} / \textbf{0.32} & \textbf{19.4} / \textbf{0.66} / \textbf{0.32} & \textbf{21.0} / \textbf{0.74} / \textbf{0.37} & \textbf{19.4} / \textbf{0.67} / \textbf{0.37} \\
    \bottomrule
    \end{tabular}
    \end{adjustbox}    
    \caption{\textbf{Quantitative Comparison of Rendering Performance} on the Matterport3D Dataset. Our method shows higher rendering quality of our method, IM360,across all scenes, achieving a 3.5 PSNR over the TexRecon \cite{waechter2014TexRecon}. The best-performing algorithms for each metric are shown in boldface. 
    }
    \label{table:mp3d-rendering-comparion}
    \vspace*{-2mm}
\end{table*}

\begin{figure}[t]
    \centering
    \includegraphics[width=1.0\linewidth]{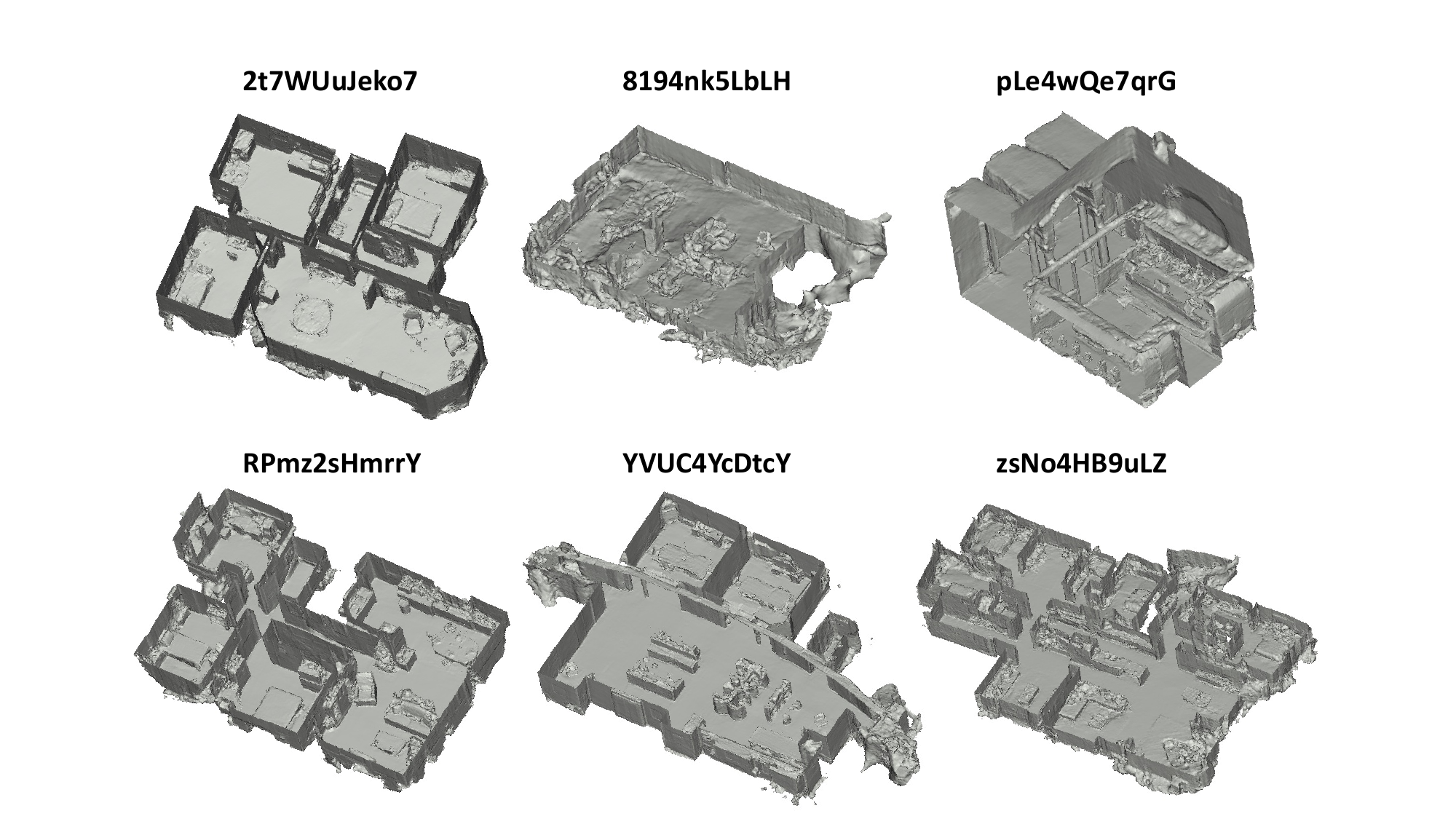}
    \caption{
    We visualize the shading meshes extracted without texture on Matterport3D. Leveraging the estimated poses from IM360, we achieve accurate surface reconstruction, demonstrating the effectiveness of our SfM method in generating decent geometric meshes from omnidirectional data.}
    \label{fig:geometry}
\end{figure}

\paragraph{Surface Reconstruction}
Recently, neural surface reconstruction methods based on volumetric rendering techniques \cite{wang2021neus, yu2022monosdf, xiao2024debsdf} have shown strong performance in dense 3D reconstruction. 
In particular, several approaches that leverage monocular geometry priors \cite{yu2022monosdf, xiao2024debsdf} demonstrate robustness in handling large textureless regions and sparse input views. 
Therefore, we adopt DebSDF \cite{xiao2024debsdf} to estimate the signed distance function and apply the marching cubes algorithm \cite{marchingcube} to convert the neural implicit representation into triangle meshes. Figure \ref{fig:geometry} presents shading meshes extracted without texture, the most effective way to visualize geometric differences. This highlights the effectiveness of our SfM method in producing decent geometric meshes from omnidirectional data.  

\paragraph{Texture Map Optimization}
We now evaluate the results of our view synthesis method by comparing it against four different approaches: 1) \textbf{ZipNeRF} \cite{barron2023zip}, the state-of-the-art neural radiance field method; 2) \textbf{3DGS} \cite{kerbl20233d}, which employs 3D Gaussian splats rendered through rasterization; 3) \textbf{SparseGS} \cite{xiong2023sparsegs}, designed to address sparse-view challenges in 3D Gaussian splatting; and 4) \textbf{TexRecon} \cite{waechter2014TexRecon}, a classical texture mapping method for large-scale scenes. Table \ref{table:mp3d-rendering-comparion} reports the rendering quality with the PSNR, SSIM \cite{SSIM}, and LPIPS \cite{LPIPS} on Matterport3D dataset. We consistently observe that neural rendering methods, including NeRF \cite{barron2023zip} and 3DGS \cite{kerbl20233d}, perform inferior to traditional texture reconstruction techniques \cite{waechter2014TexRecon}. The significant discrepancy between training PSNR and novel view PSNR indicates that this gap arises due to the sparsity of the scanned images. To address this, we evaluated the state-of-the-art SparseGS \cite{xiong2023sparsegs} approach, specifically designed for sparse-view scenarios. While SparseGS \cite{xiong2023sparsegs} achieves a 1 PSNR improvement in rendering quality, this gain indicates that sparse scanning remains a significant challenge. Our method achieves improved rendering quality across all scenes, demonstrating a 3.5 PSNR increase over the TexRecon \cite{waechter2014TexRecon} method.

\begin{figure*}[h]
    \centering
    \includegraphics[width=0.97\linewidth]{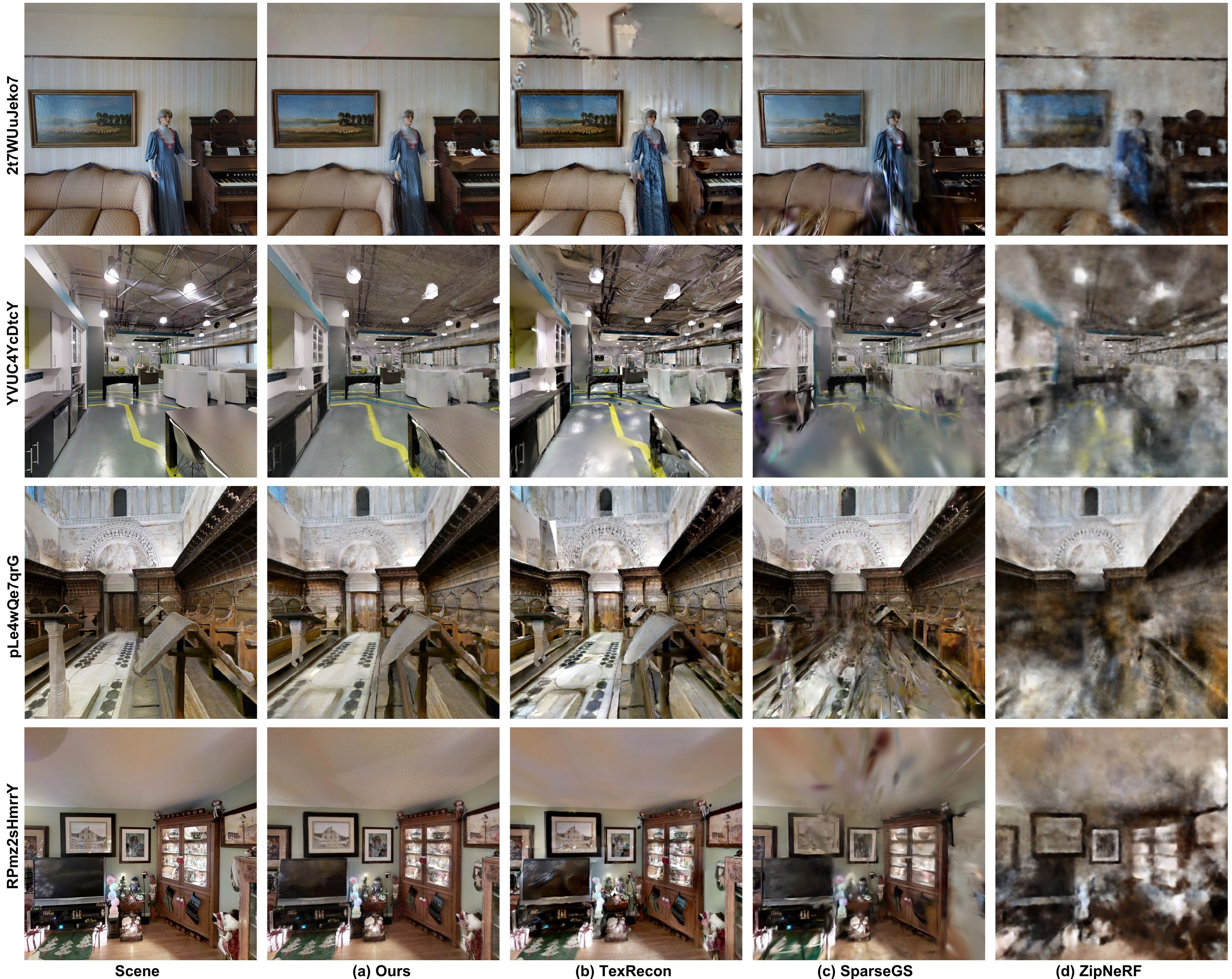}
    \caption{\textbf{Qualitative Comparisons with Existing Methods.} We visually compare our proposed approach with TexRecon \cite{waechter2014TexRecon}, SparseGS \cite{xiong2023sparsegs}, and ZipNeRF \cite{barron2023zip}. Our method can render high frequency details and results in lower noise.}
    \label{fig:rendering-comparions-matterport}
\end{figure*}

In Fig. \ref{fig:rendering-comparions-matterport}, we visually compare our method to several recent rendering approaches: TexRecon \cite{waechter2014TexRecon}, SparseGS \cite{xiong2023sparsegs}, and ZipNeRF \cite{barron2023zip}. We visualize only SparseGS instead of 3DGS, as both methods employ similar rendering techniques, but SparseGS demonstrates superior rendering quality in sparsely scanned indoor environments. 
TexRecon shows the visual seams between face texture and color misalignment.  ZipNeRF tends to generate ambiguous pixels, resulting in severe artifacts. SparseGS contains obvious floating artifacts in ground and floor regions. In contrast, our method produces much higher quality textures and consistently demonstrate improvement across various scenes. 

\begin{figure}[t]
    \centering
    \includegraphics[width=0.9\linewidth]{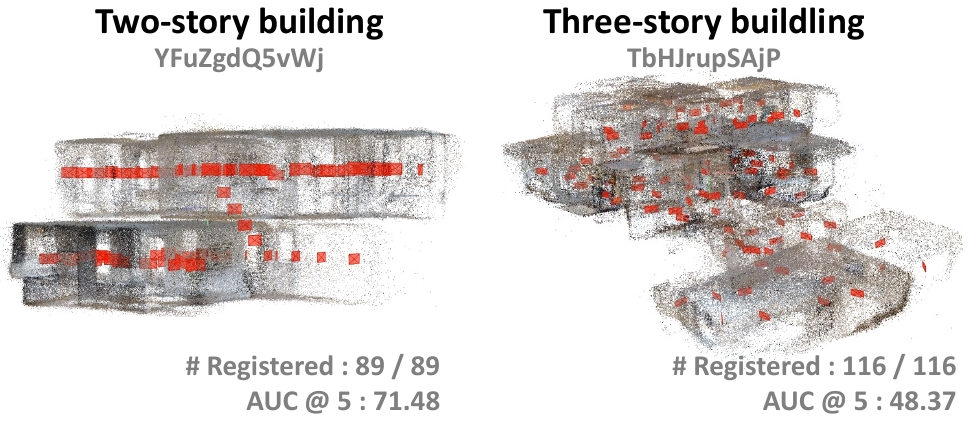}
    \vspace*{-3mm}
    \caption{Localization results in multi-floor scenes of Matterport3D.
    Multi-floor scenes pose significant challenges for visual localization due to narrow corridors and staircases connecting different floors.
    Despite these difficulties, IM360 demonstrates the ability to estimate poses, ensuring robust localization in large-scale indoor environments.}
    \label{fig:multifloor}
    \vspace*{-3mm}
\end{figure}
\subsection{Additional Experiments and Ablation Study}
We observed that when initial poses are determined using two-view geometry without applying the method proposed by Solarte et al.~\cite{solarte2021robust}, registration performance significantly deteriorates, resulting in fragmented SfM results. 
Furthermore, as shown in Table \ref{table:mp3d} and \ref{table:stfd}, by modifying the feature matching in COLMAP-based methods and comparing the registration and pose accuracy, we confirmed that our approach demonstrates superior robustness.
We also evaluate our SfM method on multi-floor scenes, as illustrated in Fig. \ref{fig:multifloor}.
Our spherical SfM exhibits robust performance in multi-floor scenes, highlighting the advantages of leveraging spherical cameras within our proposed pipeline for large-scale indoor environments.

In table \ref{table:mp3d-rendering-comparion}, IM360* only finetunes diffuse texture without utilizing specular features. Without texture optimization stage outlined in Sec \ref{sec:3_3}, our rendering quality is equivalent to that of TexRecon \cite{waechter2014TexRecon}. Although we apply texture optimization soley to the diffuse texture map, our method achieves a substantial improvement in rendering quality, with a 2.7 PSNR increase. Additionally, our final approach, which combines both diffuse and specular colors, provides a further 0.8 PSNR gain by accounting for view-dependent components.

\section{Conclusion, Limitations and Future Work}

We introduce a novel framework for textured mesh reconstruction for omnidirectional cameras in sparsely scanned large-scale indoor environments. 
To tackle the challenges posed by textureless regions and sparse viewpoints, we propose a spherical Structure-from-Motion (SfM) that incorporates dense feature matching and integrates spherical camera models.
Additionally, we reconstruct texture maps and leverage differentiable rendering to enhance its quality, improving the rendering performance in sparse-view scenarios. 
One limitation of our approach is its reliance on human annotation for determining matching pairs.
While 360\deg\ image scanning scenarios typically involve a small number of captures, making manual annotation feasible, integrating an automated image retrieval system would be a more scalable solution.
For geometric reconstruction, we employ cubemap projection to transform 360\deg\ images into perspective views and utilize state-of-the-art volumetric rendering methods. 
However, this approach has limited capacity for reconstruction, as the quality deteriorates when the scene exceeds a certain scale.
In future work, we aim to further refine geometric reconstruction and enhance the overall pipeline for large-scale 360\deg\ mapping.

\clearpage
\setcounter{page}{1}
\maketitlesupplementary

In this supplementary material, we provide additional experimental results, highlighting the superior performance of our method compared to existing approaches.
Appendix \ref{realworld} shows the qualitative results of our custom dataset to demonstrate the practicality of our method.
Table \ref{table:dataset_comparison} present the sensor configurations of publicly available datasets containing indoor scenes.
Since Matterport3D \cite{chang2017matterport3d} and Stanford2D3D \cite{Stanford2d3d} offer sparsely scanned panoramic images in large-scale environments, our experiments primarilty focus on these two datasets.
Appendix \ref{sphericalsfm} highlights the advantages of the spherical camera model and the dense matching algorithm for indoor reconstruction. 
Appendix \ref{geometry_reconstruction} describes the details and justifies the model choices for geometry mesh reconstruction.
Appendix \ref{texture} demonstrates the effectiveness of our novel texturing method.

\begin{figure}[t]
    \centering
    \includegraphics[width=1.\linewidth]{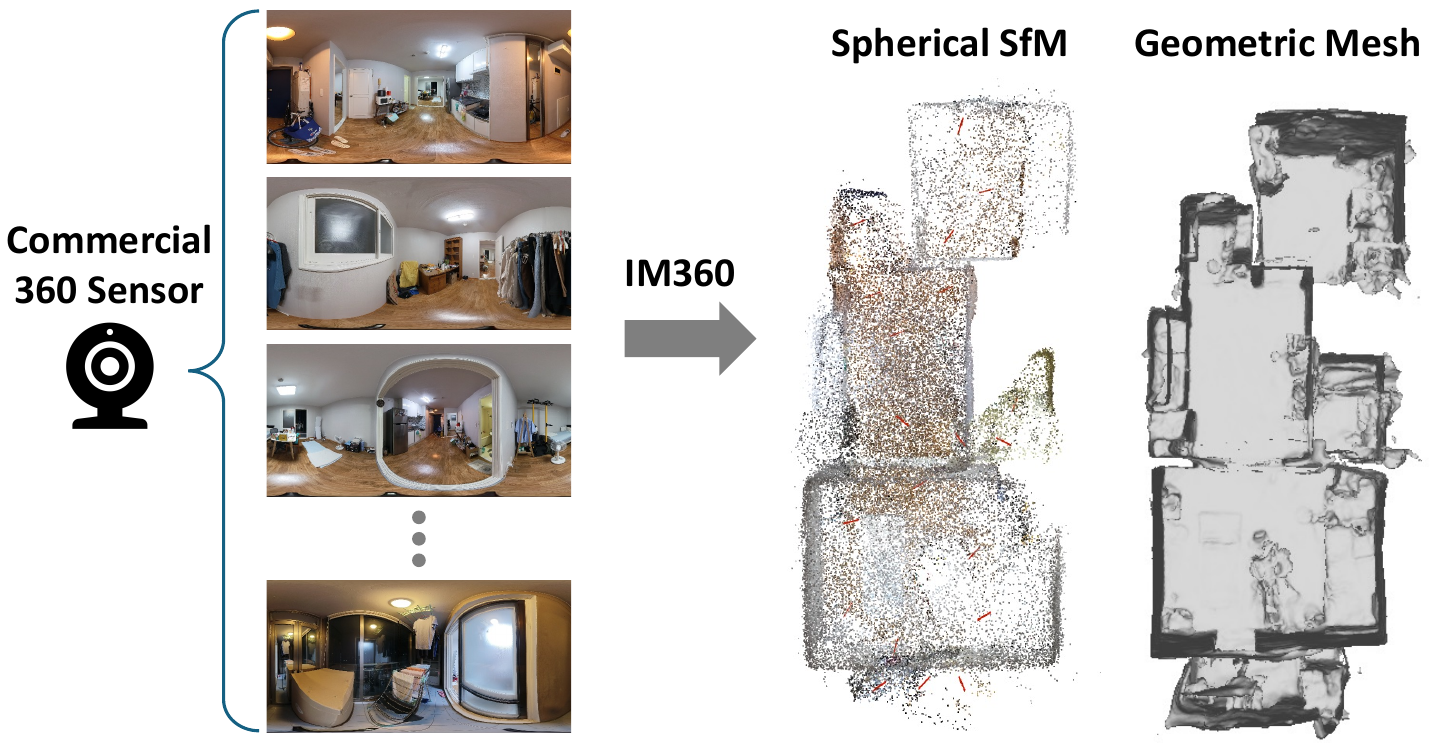}
    \caption{Qualitative results of the custom dataset. 
    We captured 20 images of a house using the commerical 360$^\circ$ sensor, Insta360.}
    \label{fig:custom}
\end{figure}

\begin{figure}[t]
    \centering
    \includegraphics[width=0.9\linewidth]{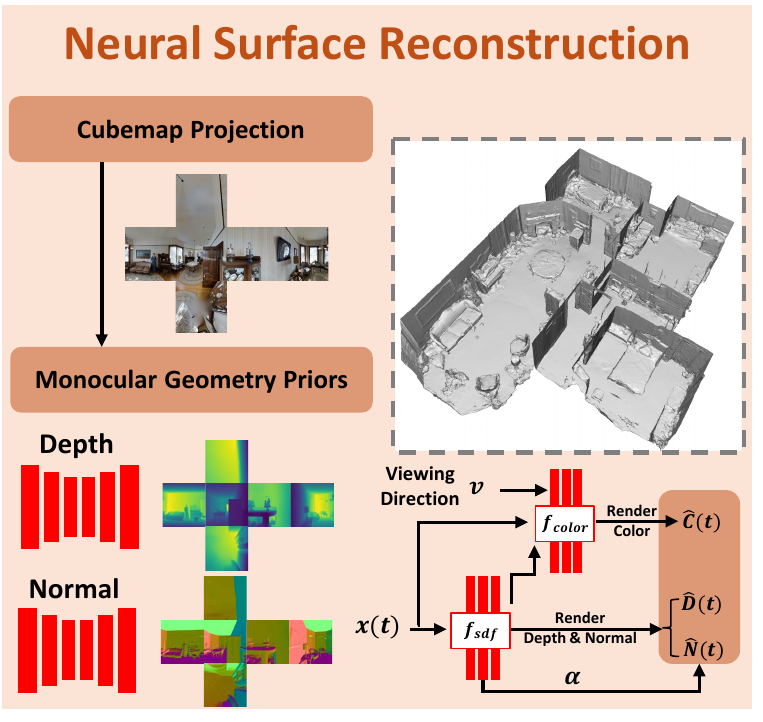}
    \vspace{-3mm}
    \caption{Overview of Neural Surface Reconstruction. In this work, we follow DebSDF \cite{xiao2024debsdf}, which estimates an implicit surface representation by utilizing volumetric rendering and monocular geometric priors. 
    This approach refines the underlying structure and enhances surface details, leading to more accurate 3D reconstructions compared to other recent methods.}
    \label{fig:debsdf}
    \vspace{-5mm}
\end{figure}

\begin{table*}[t]
    \centering
    \resizebox{\linewidth}{!}{\normalsize
    \begin{tabular}{lccccc}
    \toprule
    Dataset & Environment & Floor Space (m\(^2\)) & \# Perspective Images & \# Panoramic Images & Annotation  \\ 
    \midrule
    Stanford2D3D \cite{Stanford2d3d}  & 6 large indoor areas & 6020  & 70,496 & 1,413 & Laser Scanner \\
    Matterport3D \cite {chang2017matterport3d} & 2056 rooms (90 scenes) & 46,561 & 194,400 & 10,800 & Laser Scanner \\
    ScanNet  \cite{dai2017scannet} & 707 small rooms & 34,453 & 2,492,518 & - & RGB-D \\
    7 Scenes \cite{shotton2013seven7scenes} & 7 small rooms & - & 43,000 & - & RGB-D \\
    TUM Indoor \cite{Huitl2012tumindoor} & building (7 floors) & 16,341  & 48,974 & - & Laser Scanner\\
    TUM-LSI \cite{walch2017tumlsi} & building (5 floors) & 5,575 & 1,095 & - & Laser Scanner\\
    InLoc \cite{taira2018inloc} & building (5 floors) & 10,370 & 10328 & - & Laser Scanner / RGB-D \\
    Baidu \cite{sun2017baidu}  & mall & 9,179 & 2,078 & - & Laser Scanner \\
    NAVER LABS \cite{lee2021naverlabs} & mall and metro & 53,036 & 136,783 & - & Laser Scanner \\
    \bottomrule
    \end{tabular}}
    \caption{Among large-scale indoor datasets, Matterport3D \cite{chang2017matterport3d} and Stanford2D3D \cite{Stanford2d3d} provide 360$^\circ$ panoramic images, unlike other datasets that primarily rely on perspective images. Compared to perspective images, 360$^\circ$ images significantly reduce the number of captures required to cover a scene.
    However, this sparse and reduced number of images introduces various challenges for visual localization and mapping pipelines. 
    To address these challenges, we integrate spherical SfM, geometric reconstruction, and texture optimization techniques into our approach.}
    \label{table:dataset_comparison}
\end{table*}

\section{Real World Application}
\label{realworld}
To demonstrate the versatility of our proposed spherical Structure from Motion, we tested it on a custom dataset.
We directly captured 20 omnidirectional images in a 56 m$^2$ house using an Insta360 camera.
Figure \ref{fig:custom} presents all registered cameras obtained through spherical SfM along with the 3D reconstruction results.

\section{Technical Details}
\subsection{Spherical Structure from Motion}
\noindent\textbf{Spherical Dense Matching:}
We provide details on spherical dense matching. 
Dense matchers estabilsh pixel-wise correspondences and sample reliable matches using confidence scores.
Ideally, if the network is well-trained, unreliable matches are filtered by confidence, and further refined through geometric filtering during SfM. 
Furthermore, following the detector-free setting~\cite{shen2022semi,he2024detector},  
we quantize each 2D match location onto a grid to enhance the consistency of closely spaced subpixel matches. Specifically, each coordinate \( x \) is rounded to the nearest grid point using the formula,
$\left\lfloor \frac{x}{r} \right\rceil \cdot r$
where \( r \) denotes the grid cell size and \( \left\lfloor \cdot \right\rceil \) represents the rounding operator.  
This quantization step combines nearby matches into the same grid cell, effectively merging redundant or noisy matches into a single representative location.

\noindent\textbf{Spherical Two-view Geometry Estimation:}
We provide additional details on spherical two-view geometry estimation in the manuscript. 
Following Solarte \textit{et al.}~\cite{solarte2021robust}, we adopt the eight-point algorithm (8-PA) to estimate the essential matrix from corresponding points. 
Unlike the conventional 8-PA~\cite{hartley2003multiple}, which uses normalized image coordinates, we replace them with unit bearing vectors on a unit sphere via spherical projection. 
Thus, a minimum of $n \geq 8$ bearing vector correspondences is required to estimate the essential matrix.
We reformulate Equation~(2) from the manuscript as a least-squares problem:
\begin{equation} \label{eq:leastsquare}
    A[E]_{v} = 0
\end{equation}
Here, $A$ is an $n \times 9$ matrix formed by stacking the Kronecker products of corresponding bearing vectors as $A_{i} = u^{i}_{1} \otimes u^{i}_{2}$, and $[E]_v$ is a vector obtained by row-wise concatenation of the essential matrix entries. 
The solution to Equation~\ref{eq:leastsquare} is obtained using the Direct Linear Transformation (DLT) method~\cite{hartley2003multiple}, from which the essential matrix can be recovered.

\noindent\textbf{Image Pair Selection:} We do not consider image pair selection a core component of our pipeline and do not incorporate image retrieval methods such as NetVLAD, as we regard this as a separate research topic.
We intentionally avoid using automatic annotation to ensure accurate evaluation of other components (Sec3.1), especially since human annotation is relatively easy in sparse view settings.

\subsection{Geometric Reconstruction}
Following the DebSDF \cite{xiao2024debsdf}, we jointly train two MLPs using the differentiable volumetric rendering, (i) $f_{sdf}$, which represents the scene geometry as a signed distance function, and (ii) $f_{color}$, a color network. 
The training process of \cite{xiao2024debsdf} incorporates a combination of losses, including color reconstruction loss $L_{\text{rgb}} = \sum_{r \in R}||\hat{C}(r) - C(r)||_{1}$, Eikonal loss \cite{eikonal} $L_{\text{eikonal}} = \sum_{x \in \chi}(||\triangledown f_{sdf}(x)||_{2} - 1)^2$ , and depth and normal losses,
\vspace*{-1mm}
\begin{equation}\small
\begin{aligned}
    L_{\text{depth}} &= \sum_{r \in R}||(w \hat{D}(r) + q) - D(r)||^2, \\
    L_{\text{normal}} &= \sum_{r \in R} ||\hat{N}(r) - N(r)||_1 + ||1 - \hat{N}(r)^\top N(r)||_1.
\end{aligned}
\label{eq:depthnormalloss}
\end{equation}
The depth and normal losses are derived from prior geometric cues by comparing the rendered depth $\hat{D}(t)$ and normals $\hat{N}(t)$ with the corresponding prior depth $D$ and normals $N$ from Omnidata \cite{eftekhar2021omnidata}. 
Color image $\hat{C}$ is volumetrically rendered by ray marching $\hat{C} = \sum_{i \in I} \alpha_i \, C_i \, T_i$ along with $\hat{D}$ and $\hat{N}$. We then utilize the learned SDF ($f_{sdf}$ evaluated over a uniform grid) to extract a mesh $M$ using the Marching Cubes algorithm \cite{marchingcube}.

\begin{figure}[t]
    \centering
    \includegraphics[width=0.8\linewidth]{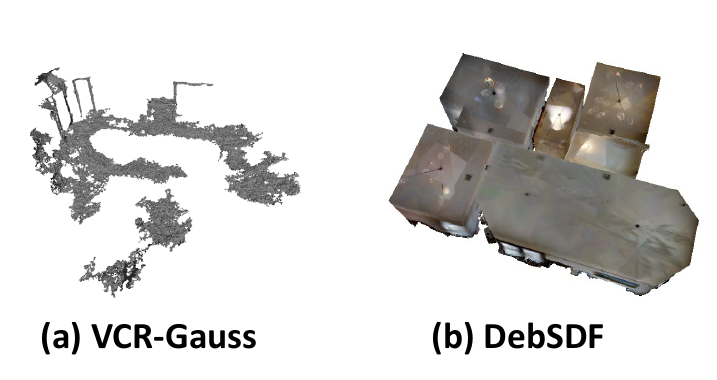}
    \vspace{-5mm}
    \caption{Examples of VCR-Gauss \cite{chen2025vcr} and DebSDF \cite{xiao2024debsdf}}
    \label{fig:vcr_gauss}
    \vspace{-5mm}
\end{figure}

\section{More Experimental Results}
\subsection{Spherical Structure from Motion}
\label{sphericalsfm}
\begin{figure*}[t]
    \centering
    \includegraphics[width=1.0\linewidth]{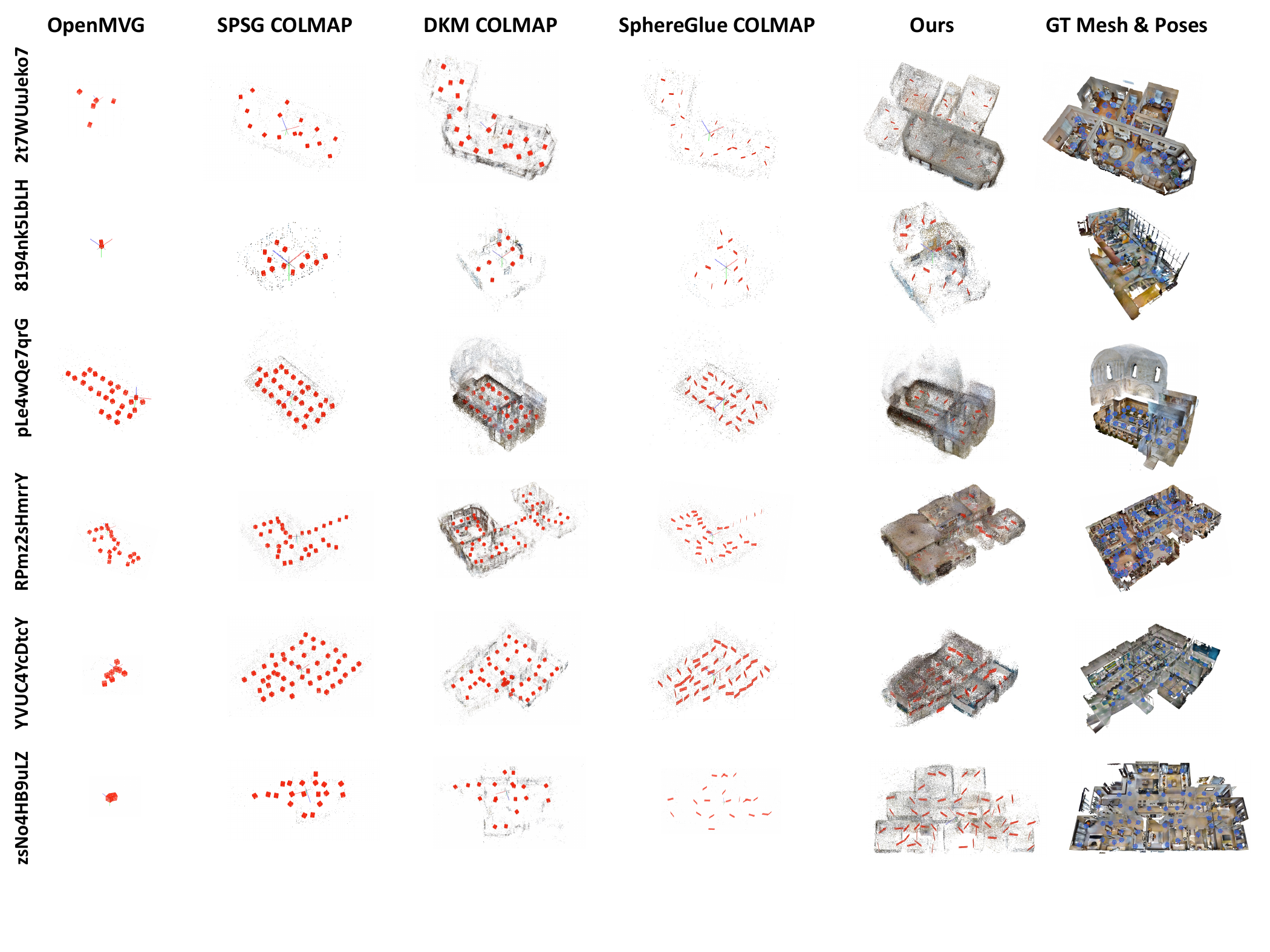}
    \caption{Qualitative Comparison of SfM results on Matterport3D.
    While other approaches failed to achieve pose registration, our method successfully estimates poses by leveraging the spherical camera model and dense matching.
    }
    \label{fig:sfm_mp3d}
\end{figure*}

\begin{figure*}[t]
    \centering
    \includegraphics[width=1.0\linewidth]{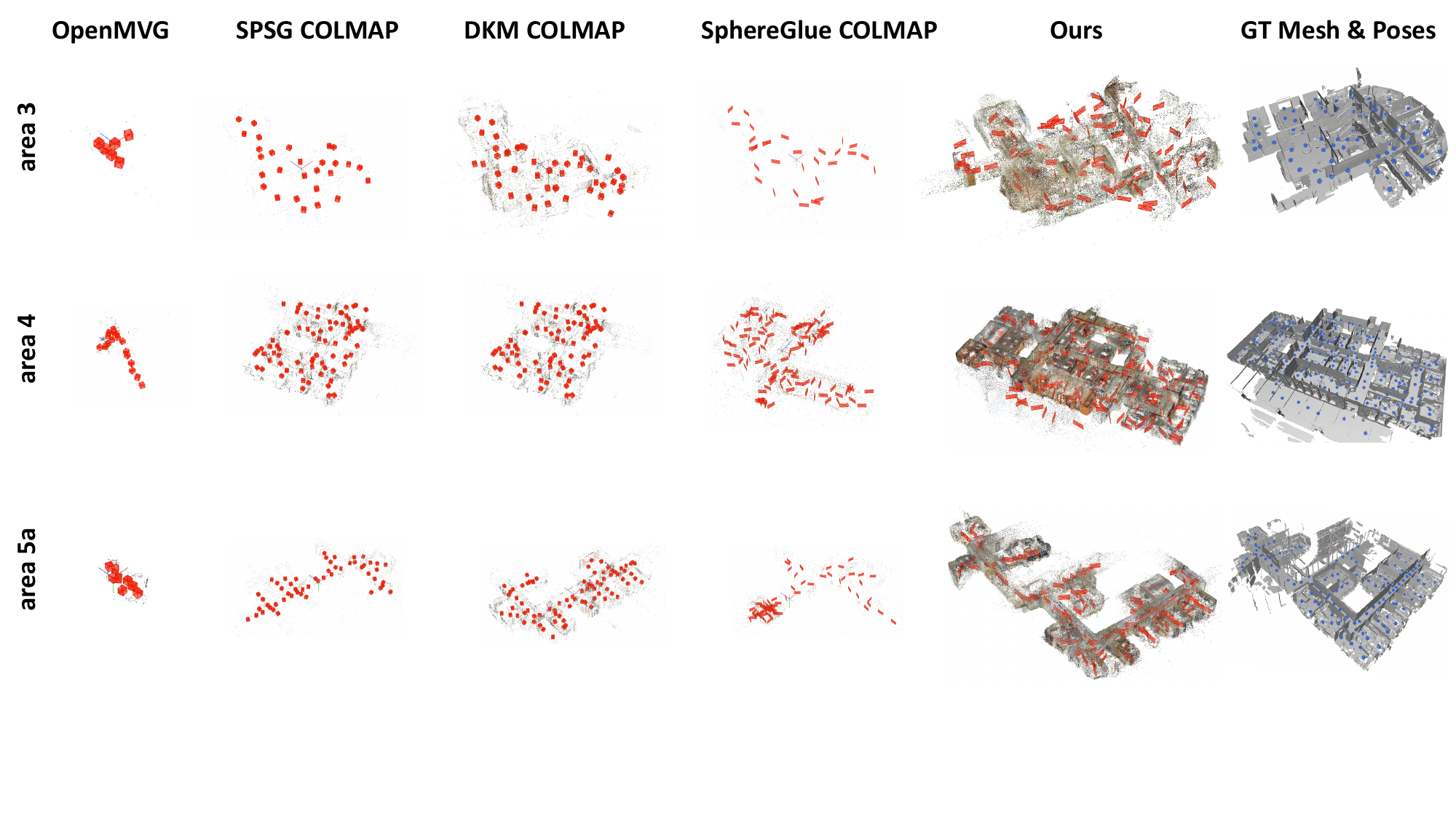}
    \caption{Qualitative Comparison of SfM results on Stanford2D3D.
    While other approaches failed to achieve pose registration, our method successfully estimates poses by leveraging the spherical camera model and dense matching.}
    \label{fig:sfm_stfd}
\end{figure*}

Due to page limitations, we present our spherical structure from motion results in the supplementary material: 1) \textbf{OpenMVG:} An open-source SfM pipeline that supports spherical camera models \cite{moulon2017openmvg}.
2) \textbf{SPSG COLMAP:} SuperPoint \cite{detone2018superpoint} and SuperGlue \cite{sarlin2020superglue} are used with cubemap and equirectangular projection.
3) \textbf{DKM COLMAP:} This method leverages DKM \cite{edstedt2023dkm} to establish dense correspondences, utilizing cubemap and equirectangular projection.
4) \textbf{SphereGlue COLMAP:} SuperPoint \cite{detone2018superpoint} with a local planar approximation \cite{eder2020tangent} and SphereGlue \cite{gava2023sphereglue} are utilized to mitigate distortion in ERP images.
The experimental results discussed in the main paper for Matterport3D \cite{chang2017matterport3d} and Stanford2D3D \cite{Stanford2d3d} are shown in Fig. \ref{fig:sfm_mp3d} and Fig. \ref{fig:sfm_stfd}, respectively.

\subsection{Geometric Reconstruction}
\label{geometry_reconstruction}

Figure \ref{fig:vcr_gauss} presents the geometry reconstruction results of VCR-Gauss \cite{chen2025vcr} for comparison. VCR-Gauss is a Gaussian Splatting-based surface reconstruction method that utilizes monocular geometry priors, similar to DebSDF \cite{xiao2024debsdf}. However, in our experiment, we completely fail to train this model, resulting in a polygonal soup.



\subsection{Texture Map Optimization}
\label{texture}
\begin{figure*}[t]
    \centering
    \includegraphics[width=1.0\linewidth]{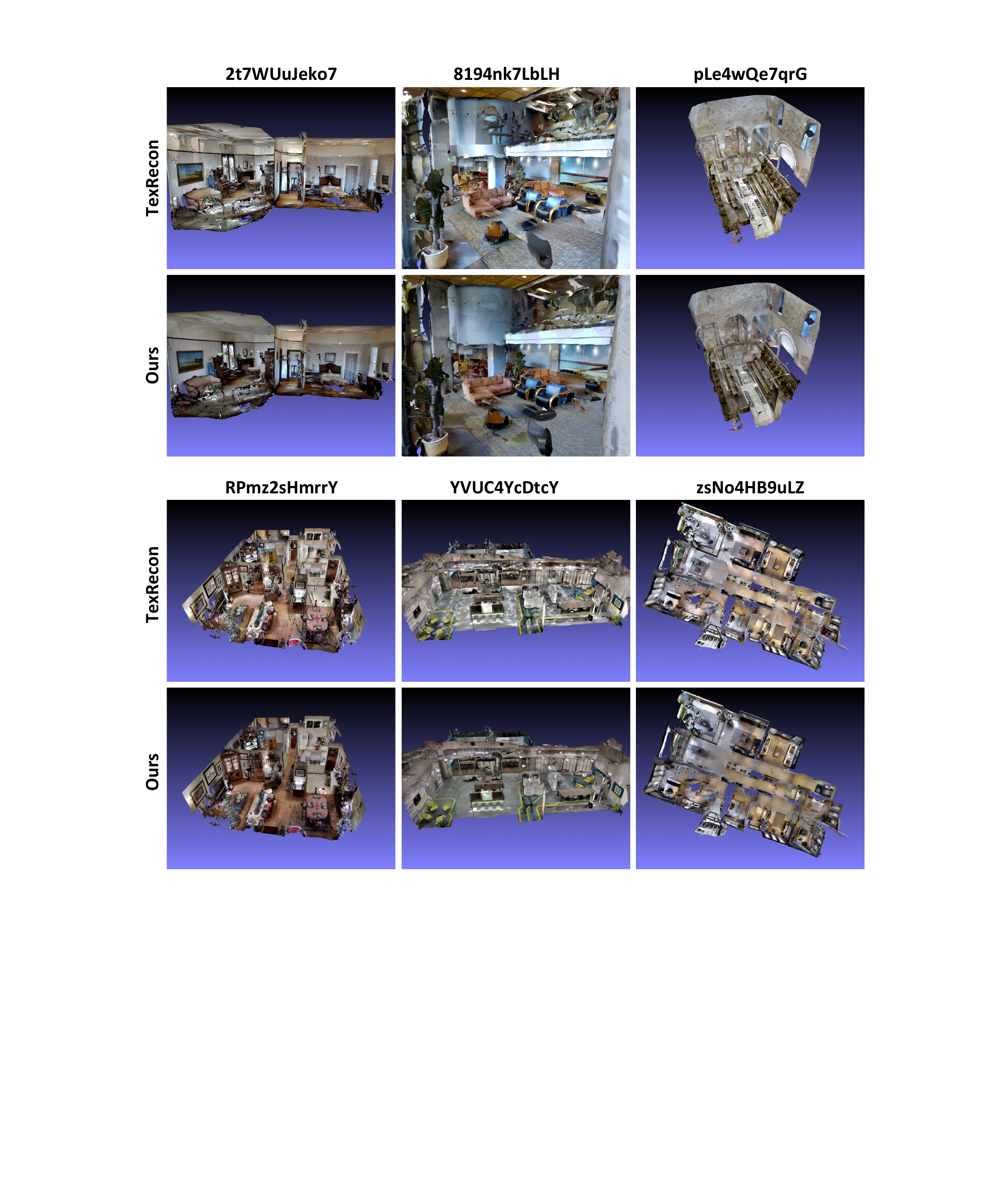}
    \caption{Qualitative Comparisons of Textured Mesh Results on Matterport3D.
    A comparison between TexRecon \cite{waechter2014TexRecon} and ours shows that our method effectively reduces noise in the texture maps, leading to improved visual quality and detail.}
    \label{fig:textured_mesh}
\end{figure*}

\begin{figure*}[t]
    \centering
    \includegraphics[width=1.0\linewidth]{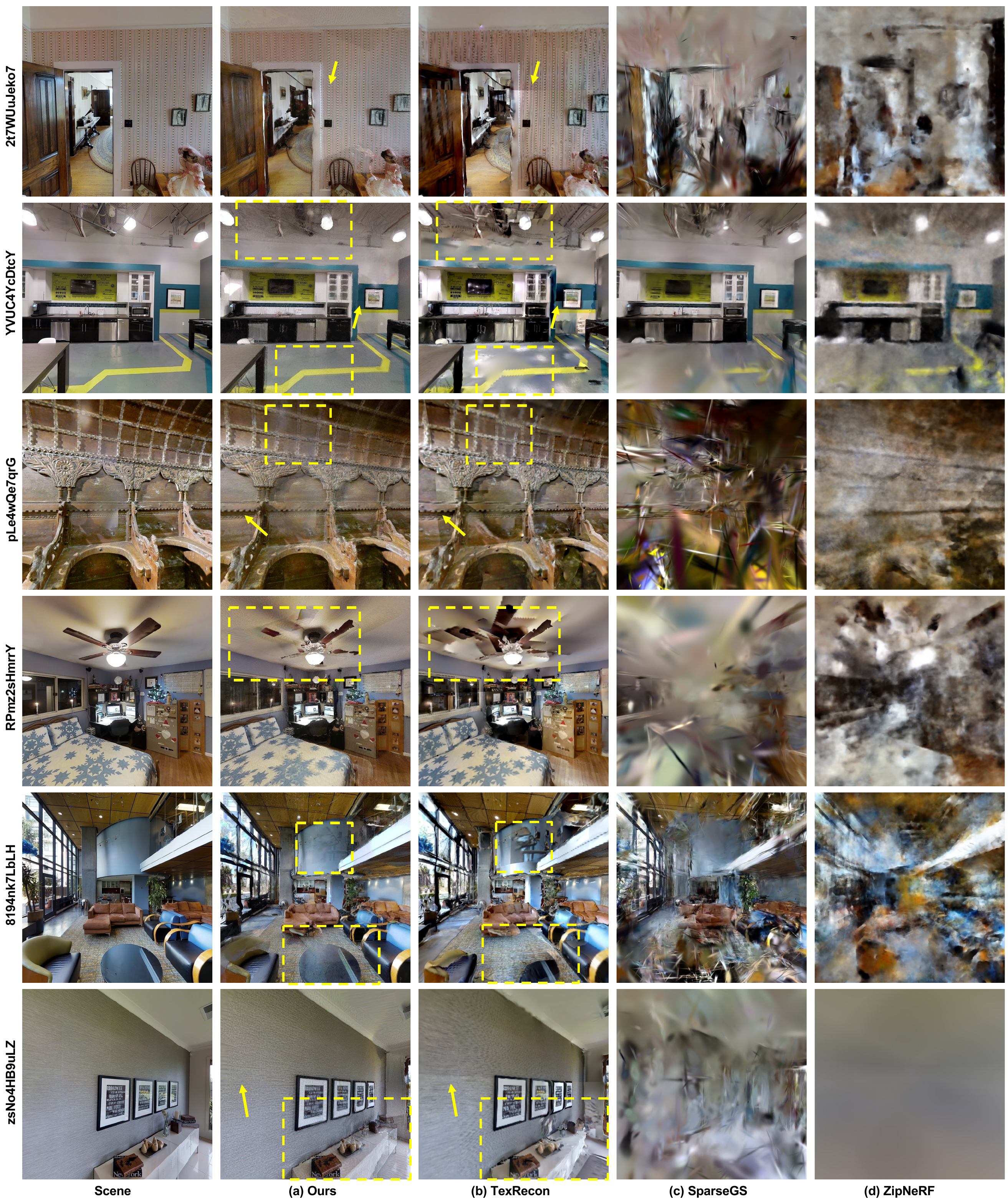}
    \caption{Qualitative Comparisons with Existing Methods. Our method can render high frequency details and results in lower noise.}
    \label{fig:render1}
\end{figure*}

\begin{figure*}[t]
    \centering
    \includegraphics[width=1.0\linewidth]{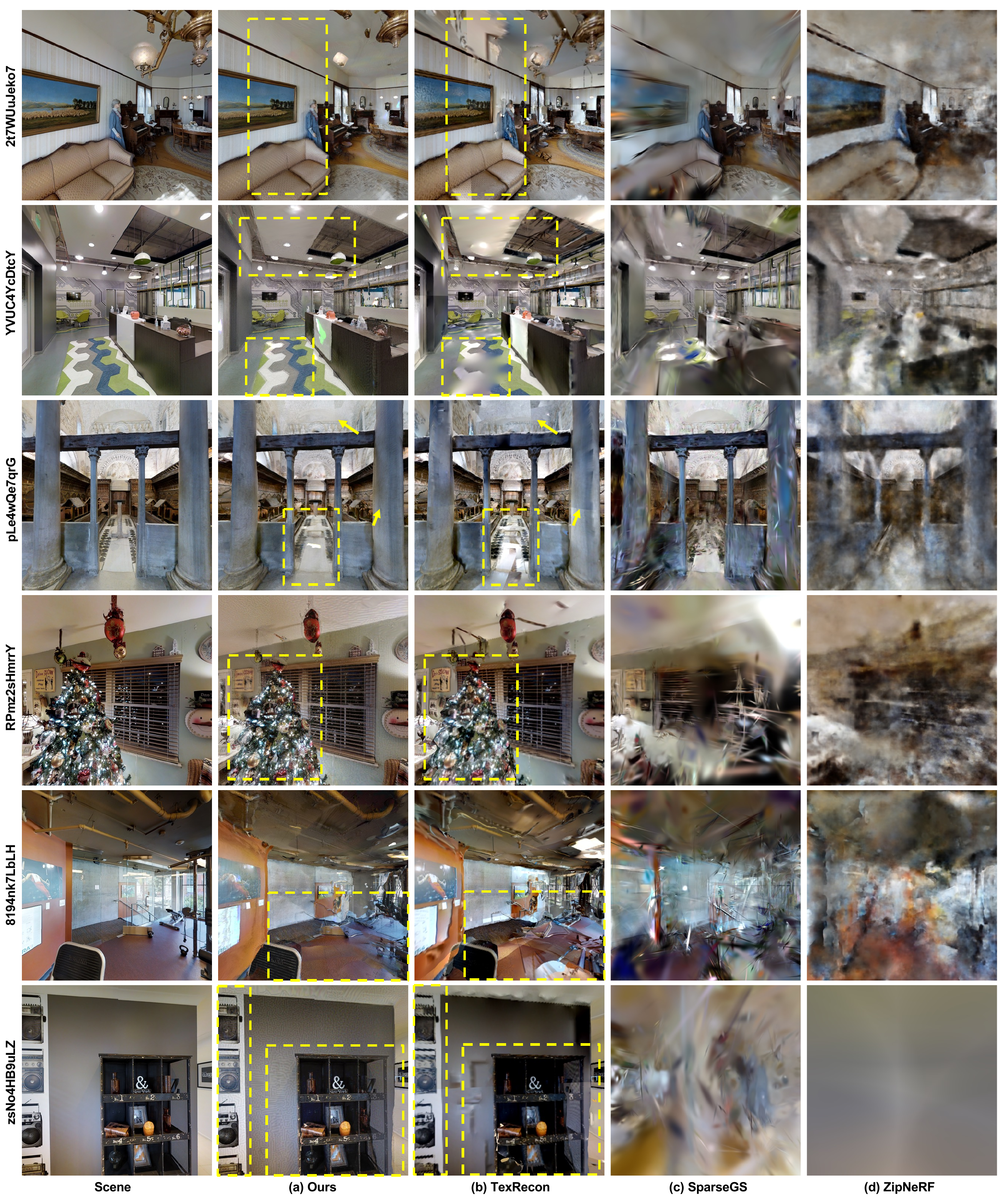}
    \caption{Qualitative Comparisons with Existing Methods. Our method can render high frequency details and results in lower noise.}
    \label{fig:render2}
\end{figure*}

\begin{figure*}[t]
    \centering
    \includegraphics[width=1.0\linewidth]{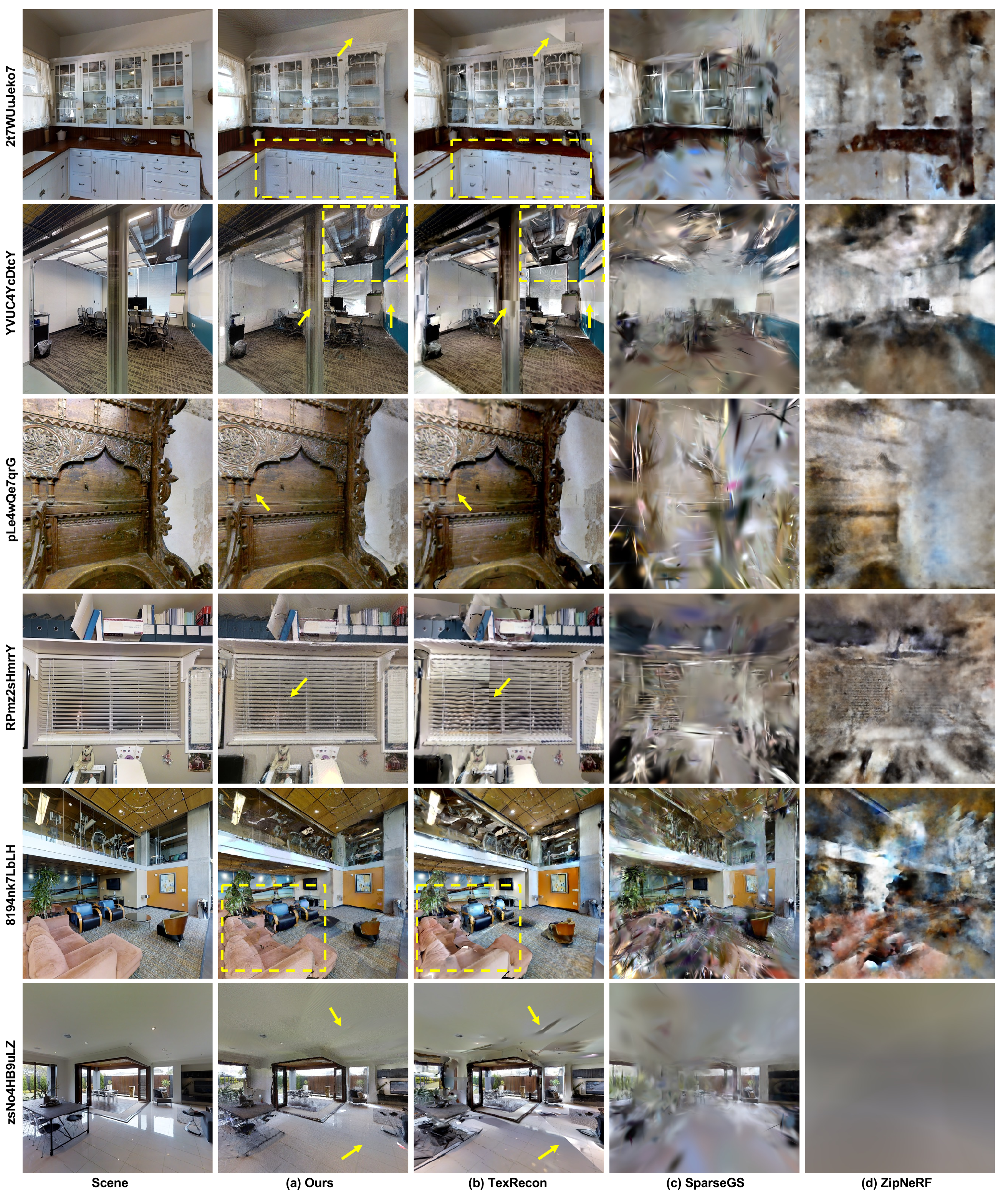}
    \caption{Qualitative Comparisons with Existing Methods. Our method can render high frequency details and results in lower noise.}
    \label{fig:render3}
\end{figure*}

\begin{figure*}[t]
    \centering
    \includegraphics[width=1.0\linewidth]{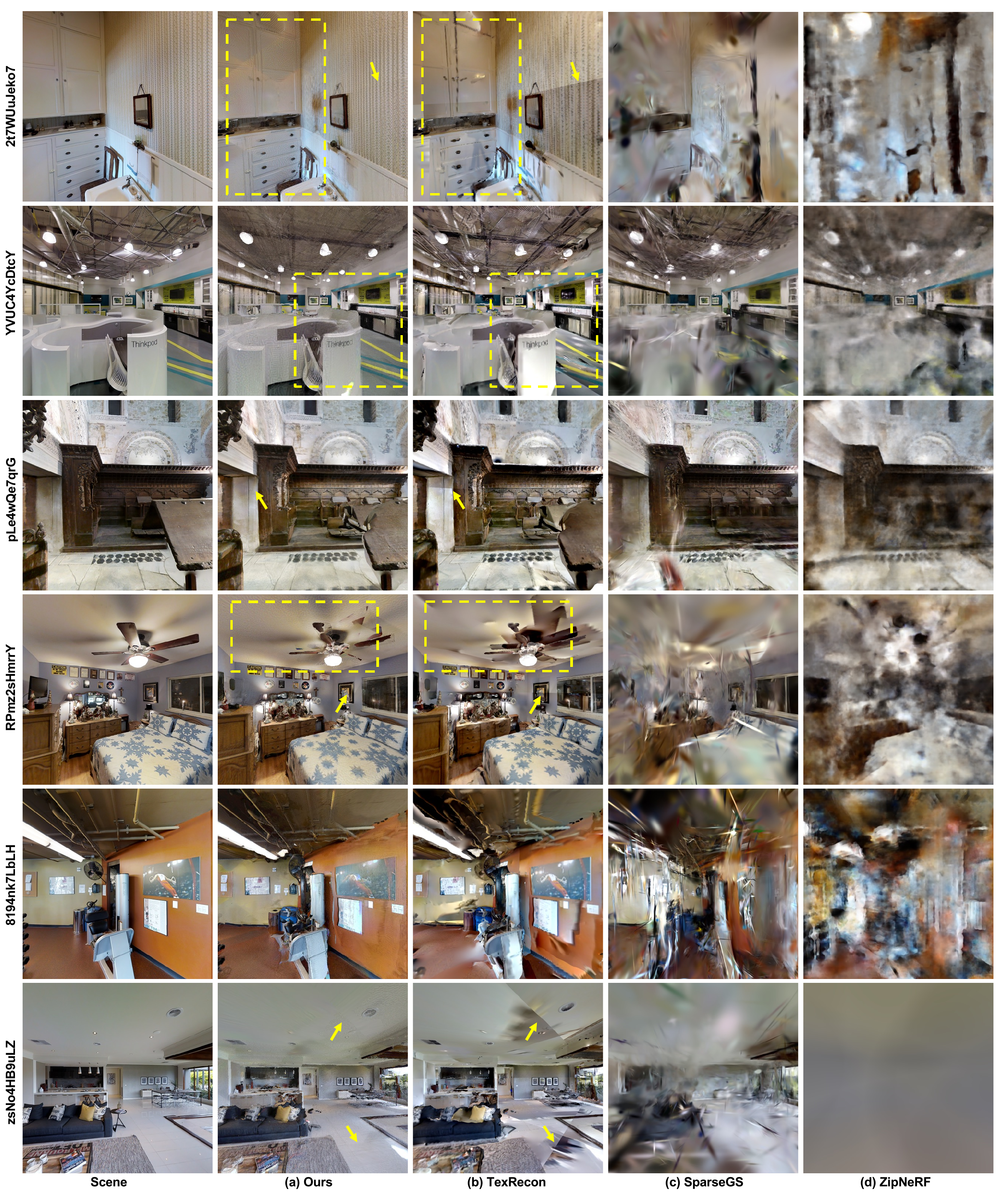}
    \caption{Qualitative Comparisons with Existing Methods. Our method can render high frequency details and results in lower noise.}
    \label{fig:render4}
\end{figure*}

We compare our method with several recent rendering approaches, including \textbf{TexRecon} \cite{waechter2014TexRecon}, \textbf{SparseGS} \cite{xiong2023sparsegs}, and \textbf{ZipNeRF} \cite{barron2023zip}. 
Our method outperforms these approaches by delivering higher frequency details and producing seamless texture maps.
The results of the textured mesh and rendering are shown in Fig. \ref{fig:textured_mesh} and Fig. \ref{fig:render1} - \ref{fig:render4}.

\subsection{Acknowledgements}
\label{ack}
This work was supported in part by ARO Grants W911NF2310046, W911NF2310352, and U.S. Army Cooperative Agreement W911NF2120076.

\clearpage

\clearpage
\twocolumn
{
    \small
    \bibliographystyle{ieeenat_fullname}
    \bibliography{main}
}
\twocolumn

\end{document}